%% file: main.tex
\documentclass[]{fairmeta}

\title{\rev{Autoregressive Diffusion for Compositional Text-to-Scene Generation}}

\author[1,2\ast]{Zhenggang Tang}
\author[1,3\ast]{Yuehao Wang}
\author[1]{Yuchen Fan}
\author[1,2]{Jun-Kun Chen}
\author[1]{Yu-Ying Yeh}
\author[1]{Kihyuk Sohn}
\author[3]{Zhangyang Wang}
\author[3]{Qixing Huang}
\author[2]{Alexander Schwing}
\author[1]{Rakesh Ranjan}
\author[1,\dagger]{Dilin Wang}
\author[1,\dagger]{Zhicheng Yan}

\affiliation[1]{Meta Reality Labs}
\affiliation[2]{University of Illinois Urbana-Champaign}
\affiliation[3]{University of Texas at Austin}
\contribution[\ast]{equal contribution}
\contribution[\dagger]{project lead}

\usepackage{hyperref}
\usepackage{url}
\usepackage[T1]{fontenc}    %
\usepackage{url}            %
\usepackage{booktabs}       %
\usepackage{amsfonts}       %
\usepackage{nicefrac}       %
\usepackage{microtype}      %
\usepackage{leftindex}
\usepackage{comment}
\usepackage[inkscapelatex=false]{svg}

\usepackage{epsfig, xspace, enumitem}
\usepackage{cases}
\usepackage{graphicx, amsmath, amssymb, caption, subcaption, multirow, overpic, textpos, pifont, adjustbox}

\usepackage[utf8x]{inputenc}
\makeatletter
\@namedef{ver@everyshi.sty}{}
\makeatother
\usepackage{pgf, tikz, pgfplots}
\usepackage{makecell}
\usepackage{pgf-pie}
\usepackage{float}
\usepackage{wrapfig}
\usepackage{colortbl}
\usepackage[title]{appendix}

\usepackage{tcolorbox}

\usepackage{listings}
\usepackage{xcolor}

\lstdefinelanguage{json}{
    basicstyle=\small\ttfamily,
    numbers=left,
    numberstyle=\tiny,
    stepnumber=1,
    numbersep=8pt,
    showstringspaces=false,
    breaklines=true,
    frame=lines,
    backgroundcolor=\color{gray!10},
    string=[s]{"}{"},
    comment=[l]{:\ "},
    morecomment=[l]{:"},
    literate=
        *{0}{{{\color{blue}0}}}{1}
         {1}{{{\color{blue}1}}}{1}
         {2}{{{\color{blue}2}}}{1}
         {3}{{{\color{blue}3}}}{1}
         {4}{{{\color{blue}4}}}{1}
         {5}{{{\color{blue}5}}}{1}
         {6}{{{\color{blue}6}}}{1}
         {7}{{{\color{blue}7}}}{1}
         {8}{{{\color{blue}8}}}{1}
         {9}{{{\color{blue}9}}}{1}
}

\newcommand{\beginsupplement}{
    \setcounter{section}{0}
    \renewcommand{\thesection}{\Alph{section}}%
    \setcounter{table}{0}
    \renewcommand{\thetable}{S\arabic{table}}%
    \setcounter{figure}{0}
    \renewcommand{\thefigure}{S\arabic{figure}}%
    \setcounter{equation}{0}
    \renewcommand{\theequation}{S\arabic{equation}}
}

\definecolor{citecolor}{HTML}{0071BC}
\definecolor{linkcolor}{HTML}{ED1C24}
\definecolor{acceptcolor}{HTML}{74C219}
\definecolor{rejectcolor}{HTML}{DE1616}
\definecolor{qcolor}{HTML}{536872}
\definecolor{demphcolor}{RGB}{100,100,100}
\definecolor{brightlavender}{rgb}{0.75, 0.58, 0.89}
\definecolor{palered}{rgb}{1.00, 0.70, 0.70}
\definecolor{palegreen}{rgb}{0.73, 0.96, 0.67}
\definecolor{paleblue}{rgb}{0.69, 0.84, 1.00}
\definecolor{paleorange}{rgb}{1.00, 0.86, 0.73}
\definecolor{palepurple}{rgb}{0.92, 0.85, 1.00}
\definecolor{paleyellow}{rgb}{1.00, 1.00, 0.50}

\renewcommand{\cite}{\citep}

\newcommand{\ard}[0]{3D-ARD\xspace}
\newcommand{\ardplus}[0]{3D-ARD+\xspace}
\newcommand{\scenesize}[0]{230K\xspace}
\newcommand{\evalscenesize}[0]{\rev{275}\xspace}

\newcommand{\rev}[1]{#1}

\makeatletter
\DeclareRobustCommand\onedot{\futurelet\@let@token\@onedot}
\def\@onedot{\ifx\@let@token.\else.\null\fi\xspace}

\def\eg{\emph{e.g}\onedot}

\makeatother

\input{sec/0_abstract}

\begin{document}

\maketitle


    

\vspace{0.3in}
\begin{center}
    \includegraphics[width=0.95\textwidth]{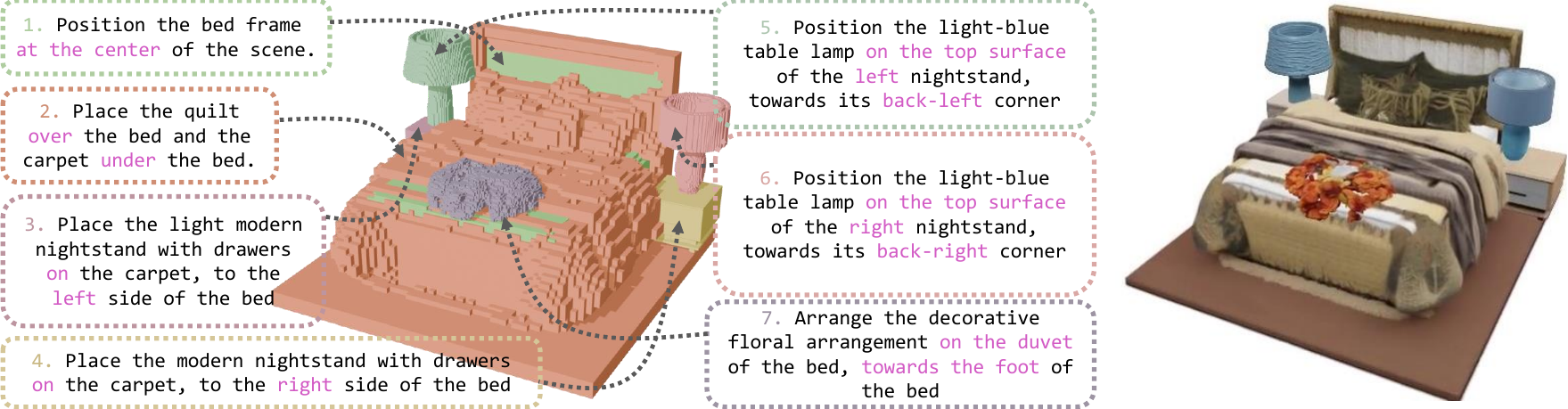}
    \captionof{figure}{We present a 3D Autoregressive Diffusion model \textbf{\ardplus}  to sequentially generate 3D objects from detailed text instructions, which not only describe the object shape and appearance, but also prescribe complex spatial relations between objects. \ardplus model generates a bedroom scene (\textbf{left}: occupancy voxel, \textbf{right}: appearance) precisely following the text instructions.}
    \label{fig:teaser}
\end{center}

\input{sec/1_intro}
\input{sec/2_related_work}

\input{sec/3_method}

\input{sec/4_experiment}

\input{sec/5_conclusion}

\input{sec/main_supp}

\clearpage
{
    \small
    \bibliographystyle{ieeenat_fullname}
    \bibliography{main}
}


\end{document}

%% file: sec/0_abstract.tex
\abstract{

Recent text-to-scene generation approaches largely reduced the manual efforts required to create 3D scenes. However, their focus is either to generate a scene layout or to generate objects, and few generate both. The generated scene layout is often simple even with LLM's help. Moreover, the generated scene is often inconsistent with the text input that  contains non-trivial descriptions of the shape, appearance, and spatial arrangement of the objects.
We present a new paradigm of sequential text-to-scene generation and propose a novel generative model for interactive scene creation. 
At the core is a 3D Autoregressive Diffusion model \ardplus, which unifies the autoregressive generation over a multimodal token sequence and diffusion generation of next-object 3D latents. To generate the next object, the model uses one autoregressive step to generate the coarse-grained 3D latents in the scene space, conditioned on both the current seen text instructions and already synthesized 3D scene. It then uses a second step to generate the 3D latents in the smaller object space, which can be decoded into fine-grained object geometry and appearance.
We curate a large dataset of \scenesize indoor scenes with paired text instructions for training. We evaluate 7B \ardplus on challenging scenes, and showcase the model can generate and place objects following non-trivial spatial layout and semantics prescribed by the text instructions. 
}

%% file: sec/1_intro.tex
\vspace{-0.3cm}
\section{Introduction}
\label{sec:intro}

The creation of immersive 3D scenes is crucial in gaming~\cite{blizzard, bhat2025cube, hu2024scenecraft, li2025worldgrow}, virtual reality~\cite{siddiqui2024meta, zhou2024dreamscene360, zhou2025il3d}, and simulation for embodied AI~\cite{yang2024holodeck, yang2024physcene, nasiriany2024robocasa, deitke2022, szot2021habitat}. This scene creation process is often interactive, where the user can compose a scene by sequentially adding objects with custom shape, appearance, and spatial arrangement. Conventional workflows~\cite{amirkhanov2025creating, schonberger2016pixelwise} often involve a time-consuming process that requires 3D artists to manually compose the scene, create detailed object geometry, and set up texture mapping. To reduce manual efforts, various text-to-scene approaches are proposed to synthesize 3D scenes from text input~\cite{yang2024holodeck, Li2024DreamScene3G, zhou2402gala3d, Yang2024SceneCraftL3, bokhovkin2025scenefactor, ling2025scenethesis, zhou2025roomcraft, wang2025hlg, yang2025sceneweaver, zhu2025imaginarium}, including models for  layout generation~\cite{tang2024diffuscene}, and object generation conditioned on layout~\cite{zhang2024towards, hu2024scenecraft, yan2024frankenstein, wu2024blockfusion}. However, fewer methods generate both~\cite{vilesov2023cg3d, fang2025ctrl}.

For scene layout generation, earlier methods generate scene layout natively, but are limited to the layout of large objects only (\eg sofa) and simple spatial relations, such as 2D layout~\cite{fang2025ctrl} and basic relations (\eg \textit{a chair next to the table})~\cite{tang2024diffuscene, vilesov2023cg3d}. They often do not handle more complex spatial relations, such as \textit{Position the table lamp on the top surface of the right nightstand, towards its back-right corner} (Figure~\ref{fig:teaser}).  More recent methods~\cite{zhang2024towards, wang2025hlg, zhou2025roomcraft, Feng2023LayoutGPTCV, fu2024anyhome, yang2024holodeck, Li2024DreamScene3G, hong2025higs} exploit LLM~\cite{hurst2024gpt, chatgpt2022} to extract scene information from text input and generate a rough layout, which, however, often deviates from the text description, does not satisfy spatial relations, and needs further heuristic optimization based on spatial constraints and object interactions (\eg Scene motif in~\cite{pun2025hsm}, refinement in AnyHome~\cite{fu2024anyhome}).  

To generate objects conditioned on the scene layout, simple methods retrieve 3D assets from external sources~\cite{tang2024diffuscene, yang2024physcene, ling2025scenethesis}, which, however, often leads to the final scene inconsistent with the textual description. Advanced methods distill 3D representation from multi-view 2D images~\cite{Yang2024SceneCraftL3, zhang2024towards, vilesov2023cg3d} or decode 3D representation from 3D latent codes generated by diffusion models~\cite{wu2024blockfusion, yan2024frankenstein}. However, generated objects are often limited to a predefined set of object categories~\cite{fang2025ctrl, bokhovkin2025scenefactor, yan2024frankenstein} or lack geometric details compared to text input~\cite{wu2024blockfusion}.

To support the interactive scene creation with detailed object shape and appearance, we propose a novel \textit{3D AutoRegressive Diffusion model (\ardplus)} to natively generate objects with diverse shapes and non-trivial spatial arrangements according to the sequential text input. When the text input prescribes fine details about the shape, appearance, and spatial relations of objects, we show that it is challenging for existing approaches, while our \ardplus model performs significantly better. The \ardplus model processes the text instructions sequentially. Each text instruction describes the shape, appearance and placement of a new object, and our \ardplus model autoregressively generates the placement, fine-grained geometry, and appearance of the new object.

Under the hood, for each text instruction, our model processes the text tokens encoded from all seen text input, and the 3D latents~\cite{trellis2024} of the current scene to predict the 3D latents of the next object in the large scene. A subsequent generation step is used to generate the 3D latents of the new object in the smaller object space, which can be decoded into fine-grained 3D Gaussians. The newly generated object is tokenized and appended to the multimodal token sequence to condition the future object generation in an autoregressive manner. The \ardplus model adopts the DiT transformer architecture~\cite{peebles2023scalable}  with causal attention between text- and 3D tokens, and unrestricted attention between 3D tokens. 

To train the \ardplus model, we curated a proprietary dataset consisting of \scenesize indoor scenes. We prompt a VLM to generate step-by-step text instructions to mimic the scene creation process. On the evaluation set, which contains \evalscenesize sets of text instructions for composing non-trivial multi-object scenes, we extensively compare our \ardplus model with several competing methods~\cite{huang2025midi, trellis2024}, and validate the \ardplus model performs significantly better in preserving the spatial relations and generating object geometry and appearance, even when object scale varies largely.

We summarize our contributions as follows:
\begin{itemize}[leftmargin=*]

\item We present a new paradigm for compositional \rev{text-to-scene} generation through an autoregressive 3D diffusion transformer, facilitating sequential object placement to synthesize complex scenes.


\item We propose a co-generation approach of layout and shape based on our autoregressive 3D diffusion modeling to achieve fine-grained generation results.

\item We develop a data pipeline to collect and process a large dataset of \scenesize indoor scenes with paired sequential text-scene data, enabling large-scale training of our 3D generation model.


\end{itemize}


\begin{figure}[t]
\vspace{-0.5cm}
\centering

\includegraphics[width=\linewidth]{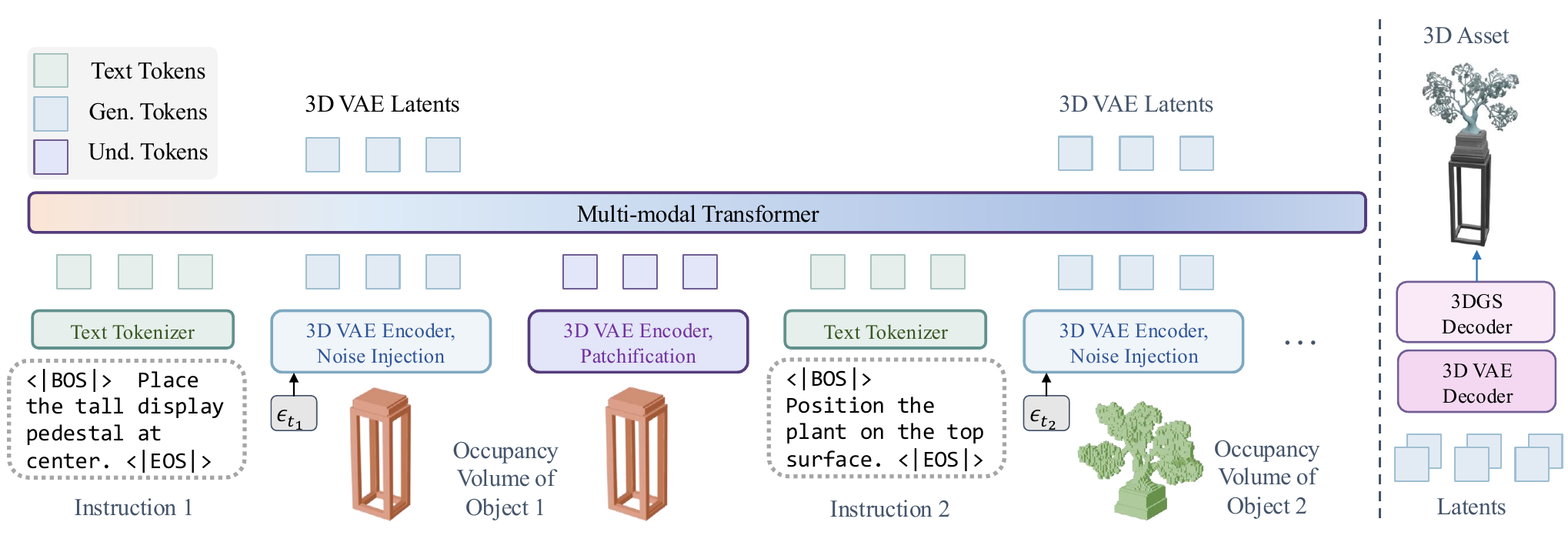}
\caption{
\textbf{Overview of \ard model for coarse-grained scene generation.} \textbf{Left}: at training time, the model takes  text tokens,  3D understanding tokens, and  noised 3D generation tokens as input, and predicts a time-dependent noise.  \textbf{Right}: at inference time, the model iteratively transforms a random noise into 3D latents, which can be decoded by 3D VAE decoder and 3D Gaussian (3DGS) decoder to generate a 3D object for each step.
}
\label{fig:overview}
\end{figure}

%% file: sec/2_related_work.tex
\section{Related Work}
\label{sec:relatedwork}

\subsection{Text-to-3D Generation}
Early Text-to-3D approaches generate 3D objects by distilling 2D diffusion priors~\cite{rombach2022high, peebles2023scalable}, without any 3D training data, including DreamFusion~\cite{Poole2022DreamFusionTU}, \rev{ProlificDreamer}~\cite{wang2023prolificdreamer}, and many others~\cite{lin2023magic3d,wang2023score,zhu2023hifa,liang2024luciddreamer}.
Multi-view diffusion models~\cite{shi2023mvdream,liu2024syncdreamer,tang2024pixel, long2024wonder3d} directly generate pose-consistent image views, which can be used to reconstruct 3D objects either via optimization~\cite{long2024wonder3d} or another multi-view transformer~\cite{tang2025mv}. Recent 3D generative models~\cite{nichol2022point,jun2023shap,trellis2024,zhang2024clay} learn to directly map text to 3D latents or explicit 3D representations but are not capable of distinguishing individual objects and preserving the spatial arrangement prescribed in the text.
It is still challenging to apply such Text-to-3D approaches to the task of interactive multi-object scene generation, where the user still needs to manually place the generated individual objects into the 3D scene.


\subsection{3D Indoor Scene Generation}
Indoor scene generation approaches often leverage LLMs~\cite{chatgpt2022,hurst2024gpt}, 2D~\cite{Podell2023SDXLIL, rombach2022high}, and 3D generative models~\cite{Poole2022DreamFusionTU,trellis2024}. While some are focused on scene layout generation, others focused on generating objects or both.
For scene layout generation, autoregressive~\cite{paschalidou2021atiss} and diffusion~\cite{tang2024diffuscene,fang2025ctrl} models are often used to generate a scene code  and 3D object attributes.
CG3D~\cite{vilesov2023cg3d} uses a diffusion model for compositional 3D object generation, but is often limited to modeling simple spatial relations.
With the tremendous advances in LLMs, many \rev{works} exploit them for scene layout generation~\cite{Feng2023LayoutGPTCV,zhang2024towards,wang2025hlg,zhou2025roomcraft, yang2024holodeck,Li2024DreamScene3G,hong2025higs,cCelen2024IDesignPL,Ran2025DirectNL}. Although LLMs improve the scene diversity, the result is often not well aligned with the text input and requires further optimizations~\cite{yang2024holodeck,cCelen2024IDesignPL,Ran2025DirectNL}. AnyHome~\cite{fu2024anyhome} prompts the LLM to convert the input text into structured representations, but still rectifies the room layout by Score Distillation Sampling~\cite{Poole2022DreamFusionTU}. HSM~\cite{pun2025hsm} uses a VLM to extract room type and objects from the text, but still needs multiple steps to generate the final layout, such as extracting support region, generating scene motif and optimizing the room layout. Our {\ardplus} model natively generates the next object conditioned on the seen text instructions and already generated objects, and thus places the next object in the scene consistent with the input.


%
To create 3D objects in the scene, \cite{tang2024diffuscene,yang2024physcene,ling2025scenethesis} retrieve 3D assets from asset libraries~\cite{Deitke2022ObjaverseAU,Deitke2023ObjaverseXLAU}, but it lacks coherence to text instruction or between objects. Instead, recent works generate 3D objects~\cite{Yang2024SceneCraftL3,zhang2024towards,vilesov2023cg3d,wu2024blockfusion,yan2024frankenstein,fang2025ctrl,bokhovkin2025scenefactor}. BlockFusion~\cite{wu2024blockfusion} and SceneFactor~\cite{bokhovkin2025scenefactor} generate 3D objects holistically as a scene, thus lacking geometric details for individual objects, and requiring further refinement.
SceneCraft~\cite{Yang2024SceneCraftL3} reconstructs 3D NeRF representation from multi-view images. CG3D~\cite{vilesov2023cg3d}, DreamScene~\cite{Li2024DreamScene3G} generate 3D objects with variants of score distillation sampling~\cite{Poole2022DreamFusionTU}. 
In contrast, our \ardplus generates 3D latents, which can be decoded into 3D representation (\eg, 3D Gaussian) of individual objects in the scene step-by-step, thereby maintaining the fine-grained details and high fidelity.

With the rise of video diffusion models (VDMs), generating 3D scenes from images or videos has been studied. ArtiScene~\cite{Gu2025ArtiSceneLA} extracts 3D attributes from isometric scene layouts and generates 3D objects via image-to-3D model. StarGen~\cite{Zhai2025StarGenAS}, HunyuanWorld-Voyager~\cite{huang2025voyager} propose long-range, pose-controllable VDMs whose frames can be turned into 3D Gaussian splats. However, these two-stage approaches are susceptible to error accumulation, and the quality of 3D reconstruction is limited by the fidelity of the intermediate images or videos.


\subsection{Multimodal Generative Models}
Since the success of visual generation models based on the text prompt~\cite{ramesh2021zero,saharia2022photorealistic,rombach2022high,dai2023emu}, it has been studied to generate visual contents beyond the text input, such as reference images. Earlier work focused on building a specialized model for each set of reference images~\cite{gal2022image,ruiz2023dreambooth}. However, such approaches are expensive and limited to making reference to only a few concepts. 

Following the success of LLMs, multimodal generative models are introduced~\cite{hurst2024gpt,comanici2025gemini,zhou2024transfusion,team2024chameleon,shi2024lmfusion,chen2025janus,xie2025muse,pan2025transfer,he2025mars,wang2025illume,wu2025janus,deng2025emerging, zhang2026thinkstrokespixelsprocessdriven}. Compared to earlier works based on fine-tuning, multimodal generative models are fast and flexible, as they take multimodal inputs (\eg reference image) in an in-context manner to generate multimodal outputs.
Chameleon~\cite{team2024chameleon}, ILLUME~\cite{wang2025illume}, MUSE-VL~\cite{xie2025muse}, and Janus-family~\cite{wu2025janus,chen2025janus} employ a discrete image tokenizer so that both text and vision modalities can be modeled using a single autoregressive transformer.
On the other hand, Transfusion~\cite{zhou2024transfusion}, LMfusion~\cite{shi2024lmfusion}, MetaQuery~\cite{pan2025transfer} and Bagel~\cite{deng2025emerging} develop models by combining the best of both worlds, where the discrete (\eg text) token is processed using next-token prediction, while the continuous (\eg image) token is processed via diffusion. 
The proposed {\ardplus} is inspired by the latter that combines next-token prediction and diffusion for multimodal generation, but is designed to generate 3D representation directly from our autoregressive-diffusion model.


%





%% file: sec/3_method.tex
\section{Method}
\label{sec:method}


We tackle the task of sequential text-to-scene generation, and the goal is to generate individual objects in a multi-step process based on sequential text instructions, which often \rev{prescribe} the shape, appearance, functional use and spatial arrangement of the new object at each step. We do not generate walls, floors and ceilings since they can be easily generated by prior methods~\cite{infinigen2024indoors, yang2024holodeck, fu2024anyhome, pun2025hsm}.

We propose a novel 3D Autoregressive Diffusion model \textbf{\ard} for coarse-grained scene generation (Figure~\ref{fig:overview}), and further extend it into a \textbf{\ardplus} model for fine-grained scene generation (Figure~\ref{fig:refine_concept}). Both are trained using a diffusion objective~\cite{rectifiedflow2023}. We address three key challenges below. \textbf{1)}: \textit{How to build a model to generate the next object with coarse shape and appearance conditioned on the text input and the already synthesized 3D scene (if any)?} In Section~\ref{sec:3d_ard}, we present the \ard model architecture. \textbf{2)}: \textit{How to generate objects with fine geometric details and appearance when the object size varies and the scene is often much larger?} In Section~\ref{sec:obj_refine}, we present an extended \ardplus model for refining the object geometry using extra refinement steps. \textbf{3)} \textit{The lack of training data with paired scene and sequential text input}. In Section~\ref{sec:data_curate}, we introduce our data pipeline to curate a large-scale indoor scene dataset with paired text instructions.

\input{sec/3_1_model_arch}
\input{sec/3_2_object_refine}

\input{sec/3_3_data_curate}

%% file: sec/3_1_model_arch.tex
\subsection{\ard: Autoregressive 3D Diffusion}
\label{sec:3d_ard}

We denote the sequential text instructions as $\{T_t\}_{t=1}^N$ where $N$ is the total steps. The text instruction $T_t$ for generating the next object often not only describes the shape and appearance of the object, but also its spatial arrangement. An example from Figure~\ref{fig:teaser} is \textit{``Position the light-blue table lamp on the top surface of the right nightstand, towards its back-right corner.''}  Therefore, the next-object generation should be conditioned on all seen text instructions and the already synthesized 3D objects (if any). Inspired by Transfusion~\cite{zhou2024transfusion}, we build a multi-modal transformer model that simultaneously processes text tokens $X^T_t$, 3D understanding tokens $X^U_t$ and 3D generation tokens $X^G_t$ at each step $t$. 

\paragraph{Text tokens.} We tokenize the text string $T_t$ into a sequence of discrete tokens with the BAGEL tokenizer~\cite{deng2025emerging}, and use standard embedding layers to convert tokens into vectors  $\operatorname{Embed}(T_t)$ of dimension $C$. At each step, we concatenate text tokens of all seen text instructions to obtain $X^T_t$.

\paragraph{3D understanding tokens.} At training time, for each of the existing objects $\{O_{t'}\}_{t'=1}^{t - 1}$ in the step $t$, 
we take a 3D binary volume $V_{t'} \in \{0, 1\}^{M \times M \times M}$, which represents the occupancy of object $O_{t'}$ in the whole scene, and use a VAE encoder to encode it into low-resolution 3D latents $S_{t'} \in \mathbb{R}^{D \times D \times D \times C_S}$. 
To reduce the number of tokens, we process $S_{t'}$ using a patchification layer with non-overlapping patches of size 2, followed by a linear projection layer to align with the text embedding dimension $C$. The 3D understanding tokens $X^U_t$ include tokens of all existing objects. 

\paragraph{3D generation tokens.} Inspired by TRELLIS~\cite{trellis2024}, the occupancy of an object in the scene can be represented by a list of active voxels $\{p_i\}_{i=1}^L$, where $p_i$ is the position index of a voxel, and $L$ the total number of active voxels. The sparse voxels $\{p_i\}_{i=1}^L$ are converted into a dense binary 3D volume $V \in \{0, 1\}^{M \times M \times M}$, which is further encoded by a 3D VAE encoder into low-resolution 3D latents $S \in \mathbb{R}^{D \times D \times D \times C_S}$. During model training, we obtain the 3D generation tokens in the current step by applying patchification with a patch size of 1 and linearly projecting a noised version of $S$ into $X^G_t$. Note the generation tokens will be later denoised and decoded to predict the occupancy of next object in the scene.

\paragraph{Training recipe.}
\label{sec:ard_train_recipe}

We train the \ard model on a curated training dataset (Section ~\ref{sec:data_curate}) with paired scene and sequential text instructions. Over the generation steps $\{t\}$, we autoregressively predict the denoised 3D latents of individual objects, which can be decoded into a 3D occupancy volume $V$ in the scene. Unlike Transfusion~\cite{zhou2024transfusion}, which predicts both text token and image patches and thus applies objectives on all output tokens, our goal is to generate the next object in the scene. 
Therefore, we apply diffusion objective to the denoised 3D generation tokens only. Specifically, we model the 3D latents distribution using the Rectified flow model~\cite{rectifiedflow2023}, where in the forward pass a noised sample is obtained based on a time-dependent linear interpolation $ \boldsymbol{x}(s) = (1-s) \boldsymbol{x}_0 + s \boldsymbol{\epsilon}$ between a clean sample $\boldsymbol{x}_0$ (which is $S$ in our case) and a random noise $\boldsymbol{\epsilon}$. In the backward pass, the noised sample is denoised according to a time-dependent flow $\boldsymbol{v}(\boldsymbol{x}, s)$, which is approximated by the transformer backbone trained with the flow matching objective below.

\begin{equation}
\mathcal{L}(\theta)=\mathbb{E}_{s,\boldsymbol{x}_0,\boldsymbol{\epsilon}}\|\boldsymbol{v}_\theta(\boldsymbol{x}, s)-(\boldsymbol{\epsilon}-\boldsymbol{x}_0)\|^2_2
\label{eqn:CFM}
\end{equation}
where $\theta$ denotes the learnable parameters.



\begin{wrapfigure}{r}{0.5\textwidth}
    \centering
    \includegraphics[width=0.48\textwidth]{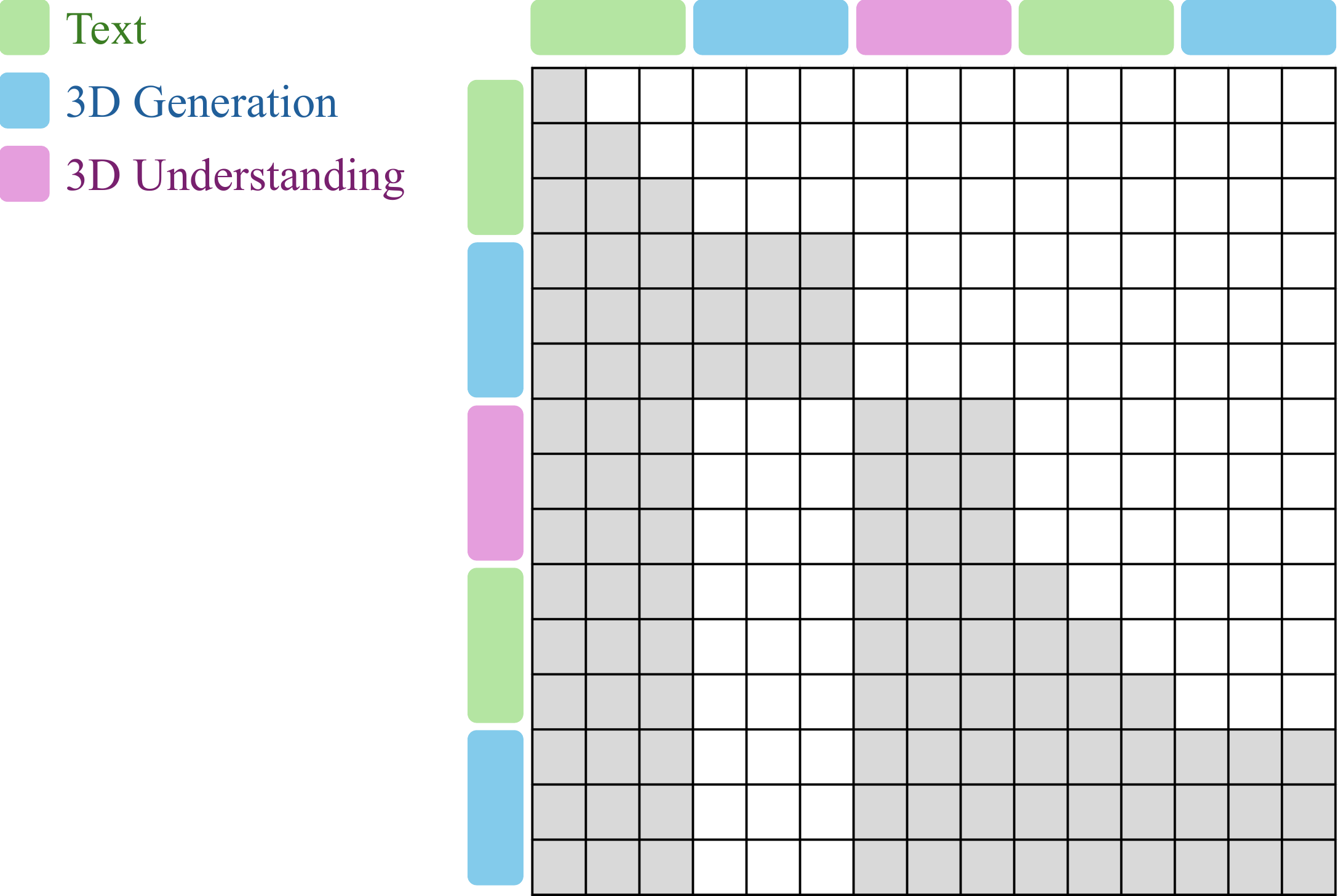}
    \caption{\textbf{The generalized causal attention used by \ard model}. Tokens along the horizontal and vertical directions are input and output tokens, respectively.}
    \label{fig:attn}
    \vspace{-0.2in}
\end{wrapfigure}

\paragraph{\ard transformer.}
The majority of the \ard model's parameters $\theta$ are with the multimodal transformer backbone, which consists of 28  blocks with self-attention~\cite{peebles2023scalable}. To explicitly reveal different types of tokens in the sequence, we insert \textit{BOS} and \textit{EOS} tokens to denote the beginning and end of text tokens. Similarly, we insert \textit{BO3D} and \textit{EO3D} tokens for 3D understanding and generation tokens. 
We implement a generalized causal attention between three types of input tokens (Figure~\ref{fig:attn}). For text tokens, it uses standard causal attention, including attention to the text tokens and 3D understanding tokens from earlier steps. 

For 3D understanding tokens, it uses the standard causal attention to text tokens, but unrestricted attention to themselves, allowing every understanding token to attend to every other understanding token in the scene. For 3D generation tokens, it uses a standard causal attention to text and 3D understanding tokens, since the generation of next object should be conditioned on all seen text and already synthesized objects. They have unrestricted attention to themselves, allowing every generation token to attend to every other generation token.

\paragraph{Model inference.}
\label{sec:ard_inference}
At inference time, for each text instruction $T_t$, we first sample a random noise $\boldsymbol{\epsilon}$ as $\boldsymbol{x}(1)$, and iteratively follow the approximate flow $\boldsymbol{v}_\theta(\boldsymbol{x}, s)$, conditioned on all past text tokens and 3D understanding tokens, to update the sample $\boldsymbol{x}(s)$ until we obtain a denoised sample $\boldsymbol{x}(0)$. We decode $\boldsymbol{x}(0)$ into a binary volume $V$ via a 3D VAE decoder to denote the occupancy of the object in the scene. To obtain the geometry and appearance of the final object, we extract the active voxels $\{p_i\}_{i=1}^L$ from $V$, iteratively denoise a randomly initialized noise by running the off-the-shelf TRELLIS 3D VAE decoder based on sparse flow transformer~\cite{trellis2024} conditioned on the current object text description to obtain the structured latent $\{(z_i, p_i)\}_{i=1}^L$, which can be decoded into 3D Gaussians using the TRELLIS 3DGS decoder (See Figure~\ref{fig:teaser}).



%% file: sec/3_2_object_refine.tex
\subsection{\ardplus: Fine-grained Scene Generation}
\label{sec:obj_refine}

\begin{figure}[t]
    \centering
    \includegraphics[width=0.75\linewidth]{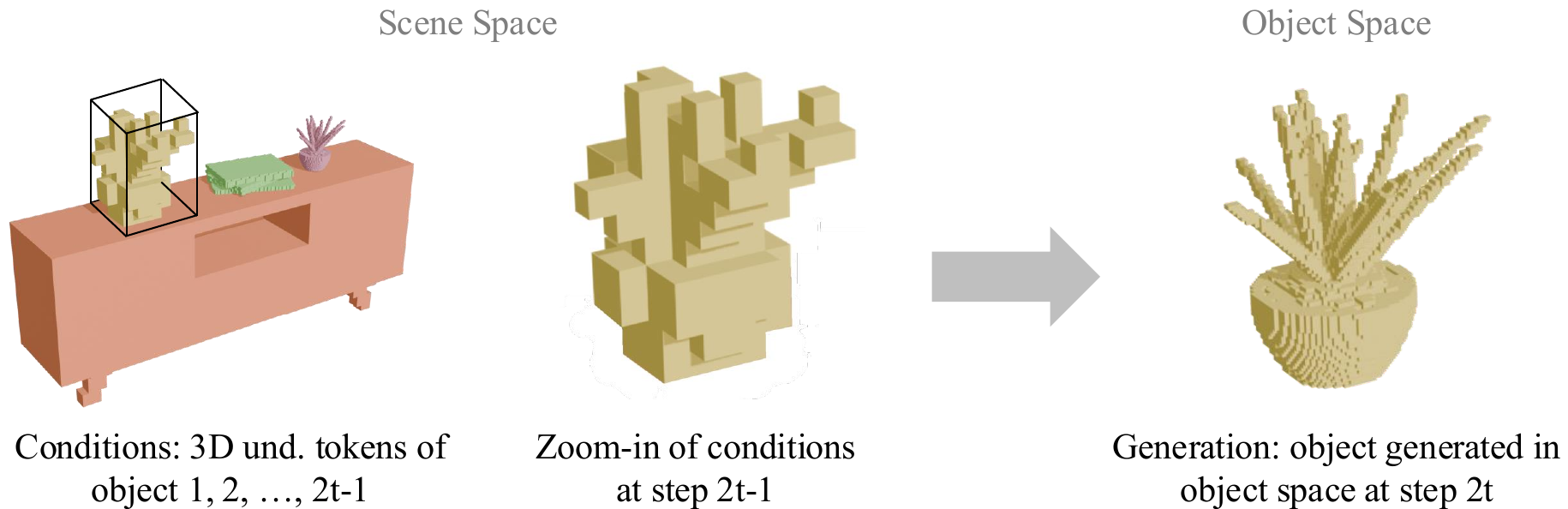}
    \caption{\textbf{Fine-grained object generation in \ardplus model.} \ardplus model generates 3D latents in the object space to obtain more fine-grained geometry. }
    \label{fig:refine_concept}
\end{figure}

Due to computational constraints, the resolution of the generated object in the scene space $V \in \{0, 1\}^{M \times M \times M}$ is limited ($M$=$64$ in our experiments), resulting in only coarse-grained object geometry for common objects (Figure~\ref{fig:ablation_refinement} Top). To address this limitation, we extend the \ard model by adding an extra step to generate 3D latents of the same resolution $D$$\times$$D$$\times D$ in the smaller object space after each current generation step, and the resulting model is referred to as \textbf{\ardplus} model. It uses a generation process of $2N$ steps for sequential text instructions $\{ T_t\}_{t=1}^N$ (Figure~\ref{fig:refine_concept}). 
Below, we present details on how to prepare tokens at even steps $\{2t\}$, where fine-grained objects are generated.

\paragraph{Text tokens.} At an even step $2t$, we take text tokens of the instruction $T_t$ already prepared at the step $2t-1 $, as well as two new special tokens \textit{BOR} and \textit{EOR}, which denote the beginning and end of refinement text, as the text tokens at the step $2t$.

\paragraph{3D understanding tokens.} At an even step $2t$, we first prepare 3D understanding tokens from all past odd steps $\{2t'-1\}_{t'=1}^{t}$ as in Section~\ref{sec:3d_ard}. For all past even steps $\{2t'\}_{t'=1}^{t-1}$, we take the groundtruth dense binary 3D volume $V_{2t'}$ in the object space, and use a 3D VAE encoder to encode it into a low-resolution 3D latents $S_{2t'}$. Similar to Section~\ref{sec:3d_ard}, latents $S_{2t'}$ are further patchified and linearly projected to reduce the token numbers and align the feature dimension.

\paragraph{3D generation tokens.} To prepare 3D generation tokens, we take the groundtruth binary occupancy volume $V$ in the object space, encode it into low-resolution 3D latents $S$ via a 3D VAE encoder, and 
apply the Rectified Flow time-dependent linear interpolation to obtain a noisy latent. As in Section~\ref{sec:ard_train_recipe}, We further patchify the latents with a patch size of 1 and apply a projection to obtain the 3D generation tokens.

\paragraph{Model training.} \ardplus model at both odd and even generation steps also attaches the conditional flow matching objective (Equation~\ref{eqn:CFM}) to the denoised 3D generation tokens to train the transformer backbone.

\paragraph{Model inference.} Similarly to the inference of the \ard model in Section~\ref{sec:ard_inference}, the 3D latents generated in the local object space are decoded into a binary occupancy volume $V$, where the active voxels with normalized coordinates are extracted to be used by the TRELLIS 3D VAE based on sparse flow transformer to generate structured latents. To put the fine occupancy volume in the scene space, we first calculate the bounding box of the coarse one and then transform the fine one accordingly.


%% file: sec/3_3_data_curate.tex
\subsection{Dataset Construction}
\label{sec:data_curate}

\begin{figure}[t]
    \centering
    \includegraphics[width=1.0\linewidth]{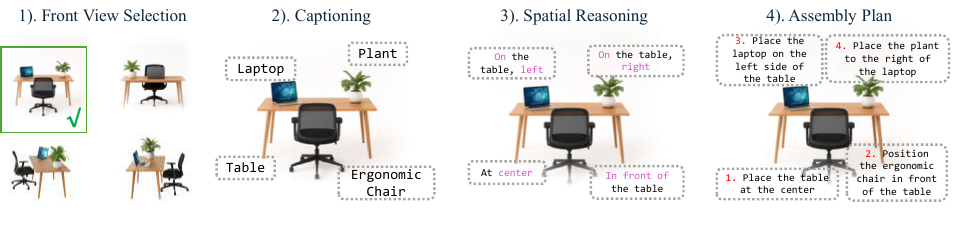}
    \vspace{-6mm}
    \caption{\textbf{Dataset construction pipeline} includes four steps: (1) Front view selection, (2) Captioning, (3) Spatial reasoning, and (4) Generating assembly plan.}
    \label{fig:data}
\end{figure}

To enable sequential scene generation, we construct a high-quality indoor dataset with step-wise assembly instructions. The raw data is from a proprietary indoor dataset, which consists of parts of a scene but lacks assembling instructions. To generate the instructions, our data pipeline consists of four main stages (Figure~\ref{fig:data}).

\paragraph{Step 1: View selection.} Given multi-view renderings of each object group, we leverage a VLM to identify the canonical front view that maximizes visual informativeness while minimizing occlusion. The model analyzes specific cues such as furniture orientation (\eg cabinet doors, bed headboards) to determine the optimal viewing angle for subsequent labeling.

\paragraph{Step 2: Object captioning.} For each object in the scene, we generate multi-granularity textual descriptions by presenting the VLM with both the complete scene view and individual object-focused views. The captions capture essential attributes including object type, material properties, color, geometric characteristics, and critically structural elements that support other objects.

\paragraph{Step 3: Spatial relationship analysis.} We incorporate 3D bounding box data (including dimensions, center positions, and spatial extents) to provide a precise geometric context for this spatial reasoning step. A top-down visualization is also generated to facilitate it.

\paragraph{Step 4: Assembly plan generation.} Combining the visual observations, object captions, geometric context, and spatial reasoning, the VLM generates a sequential assembly plan consisting of step-by-step instructions that specifies the placement order from foundational objects to dependent components. 



\label{sec:data}

%% file: sec/4_experiment.tex
\begin{figure*}[t]
\centering
\includegraphics[width=\textwidth]{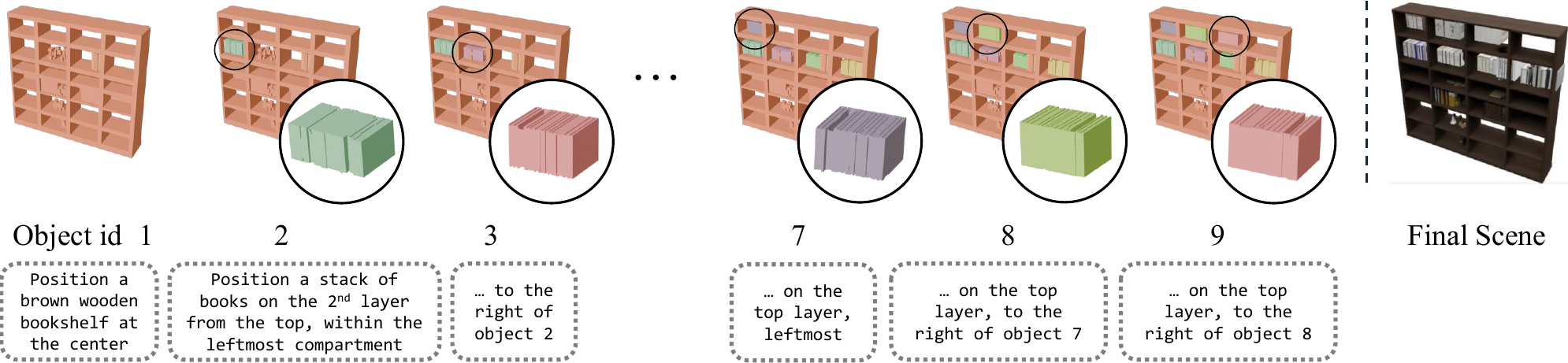}
\caption{
\textbf{An example of step-by-step generation.} Our \ardplus model closely follows the text instructions and adds a total of 8 stacks of books to different compartments in a bookshelf.
}
\label{fig:per_step}
\vspace{-0.1in}
\end{figure*}

\section{Experiments}


\input{sec/4_1_impl_details}

\input{sec/4_2_eval}
\input{sec/4_3_quali_results}
\input{sec/4_4_quant_results}

%% file: sec/4_1_impl_details.tex
\subsection{Implementation Details}
\label{sec:impl_details}
We implement our approach in PyTorch~\cite{paszke2019pytorch}. We use the dense binary occupancy volume $V \in \{0, 1\}^{M \times M \times M}$ with resolution $M$=$64$ to represent the occupancy of objects in the scene space or the smaller  object space. An \rev{off-the-shelf} 3D VAE from TRELLIS~\cite{trellis2024} is used to encode $V$ into a low-resolution continuous volume $S \in \mathbb{R}^{D \times D \times D \times C_S}$, where $D = 16$ and $ C_S = 8$. All text tokens, 3D understanding tokens, and 3D generation tokens use the feature dimension 128. 

\paragraph{\ardplus model training.} We use pre-trained multimodal transformer model from Bagel~\cite{deng2025emerging}, which contains 7B active parameters and 28 self-attention building blocks. We finetune it on our curated indoor data for 120K steps with learning rate $1e-4$ using 128 Nvidia H100 GPUs for a week. The maximum token size per sequence is 20,480.

\paragraph{\ardplus model inference.} KV pairs of text tokens for all seen text instructions are stored in the KV cache~\cite{transformer_kvcache}. KV pairs of denoised 3D generation tokens are also stored. We utilize the Euler sampler for generation, employing 50 sampling steps. The CFG coefficient~\cite{ho2022classifier} is set to 4 for text conditions and 2 for 3D condition tokens, following the configurations in \cite{deng2025emerging}.

\paragraph{Indoor training data.} Our dataset consists of 230K indoor scenes from typical room types (bedrooms, living rooms, kitchens, dining areas, bathrooms). Each scene contains 2--15 parts with diverse compositions: adjacent furniture (\eg bed with nightstands), furniture-object assemblies (\eg table with tableware), or grouped small objects. Each scene is normalized together to be within the unit bounding box. Each part has a corresponding assembly instruction (several to 40 words) describing its spatial placement. We also include a \rev{large set of 500K TRELLIS single objects} in our training data to enrich object-level diversity.

%% file: sec/4_2_eval.tex
\begin{figure}[t]
    \centering

    \scalebox{0.8}{
    \includegraphics[width=\textwidth]{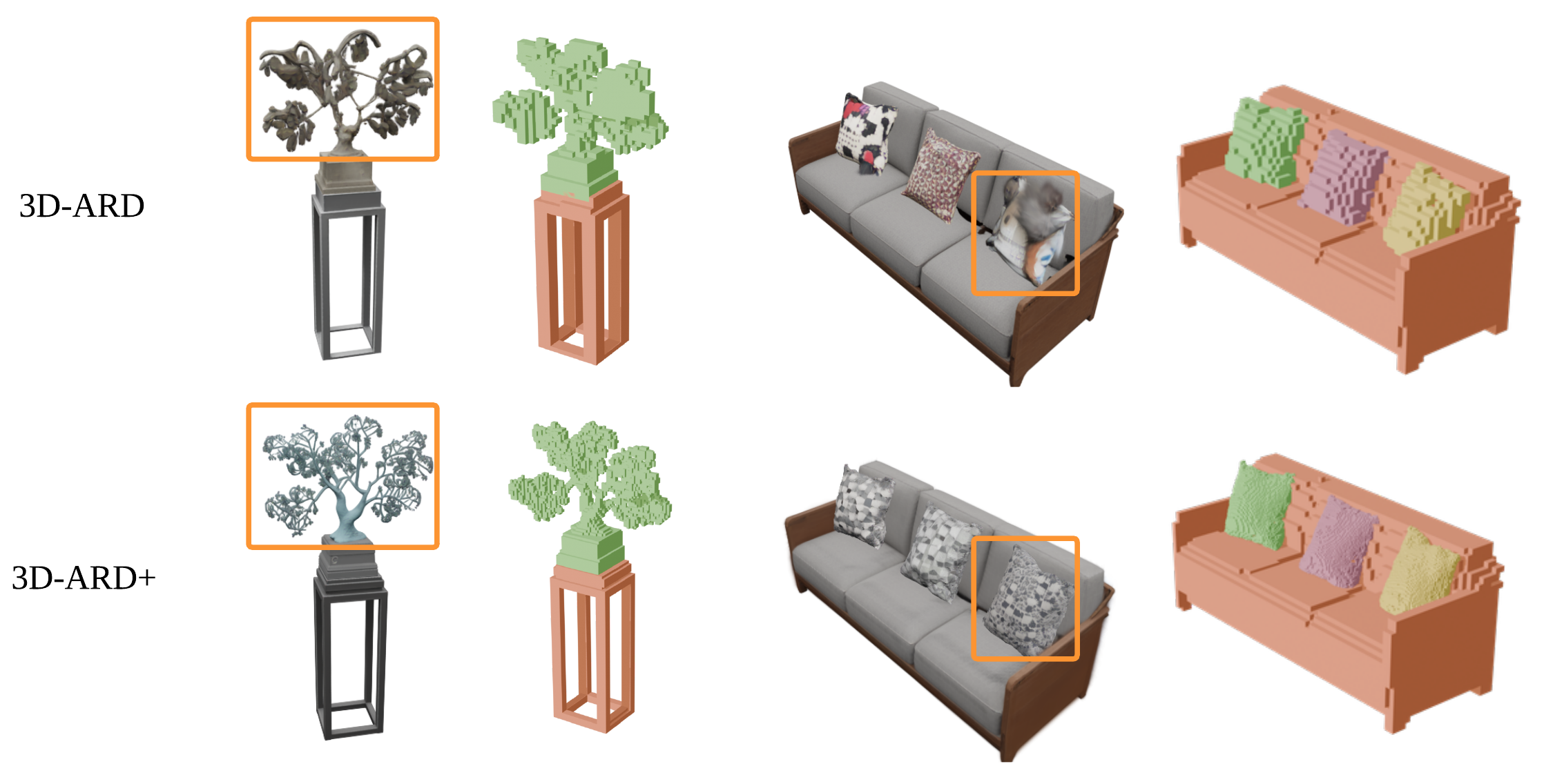}}
    \caption{\textbf{Comparing results from \ard and \ardplus models}.  Top: coarse-grained; Bottom: fine-grained.
    }%
    \label{fig:ablation_refinement}
    \vspace{-0.15in}
\end{figure}

\begin{figure}[t]
    \centering
    \includegraphics[width=0.9\linewidth]{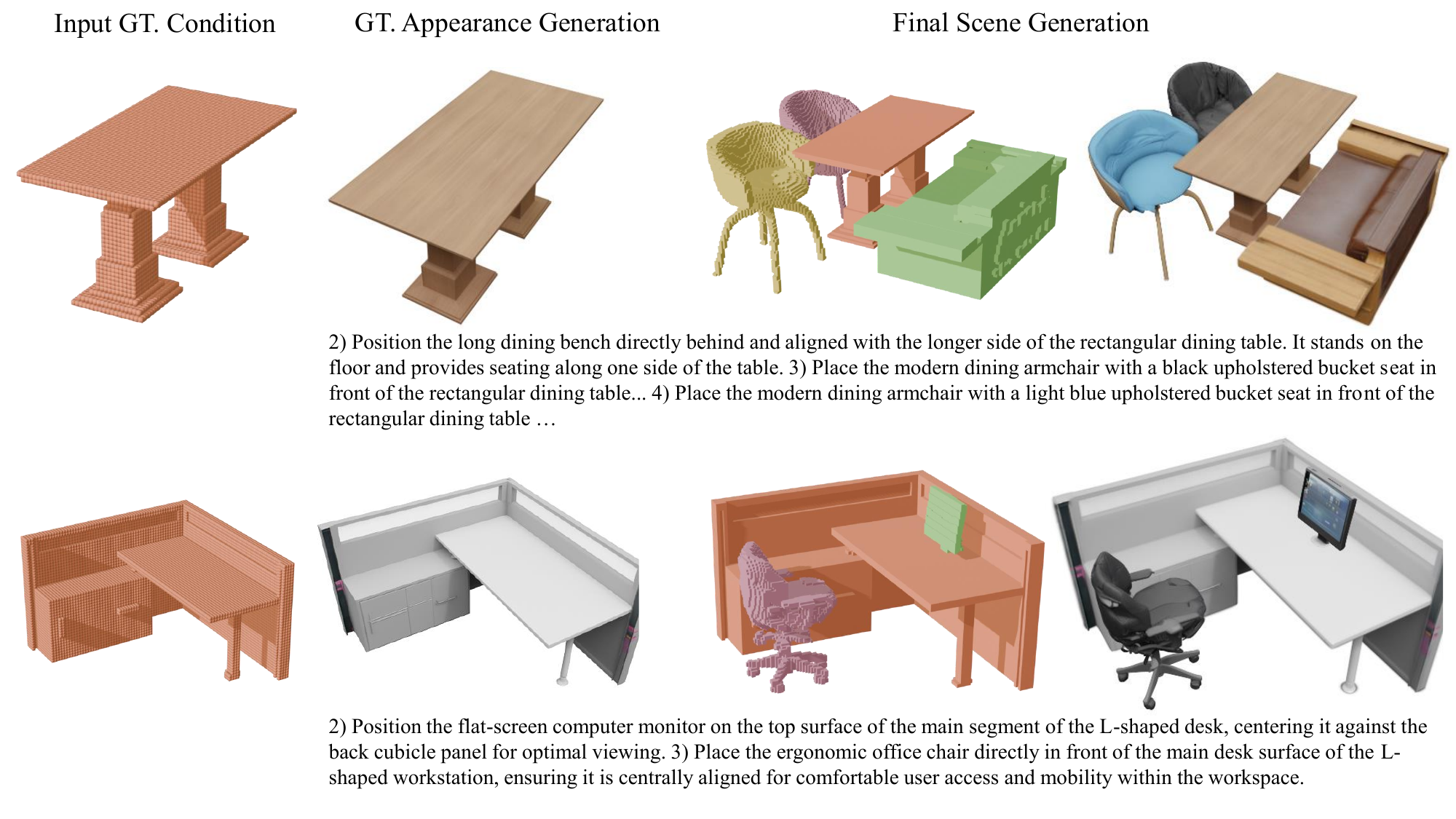}
    \caption{Conditioned next-object generation. Given the ground-truth fundamental object (table or desk) and subsequent textual instructions, our model generates subsequent objects properly.}
    \vspace{-2mm}
    \label{fig:gt_cond_gen}
\end{figure}

\begin{figure}[t]
\centering
\includegraphics[width=\linewidth]{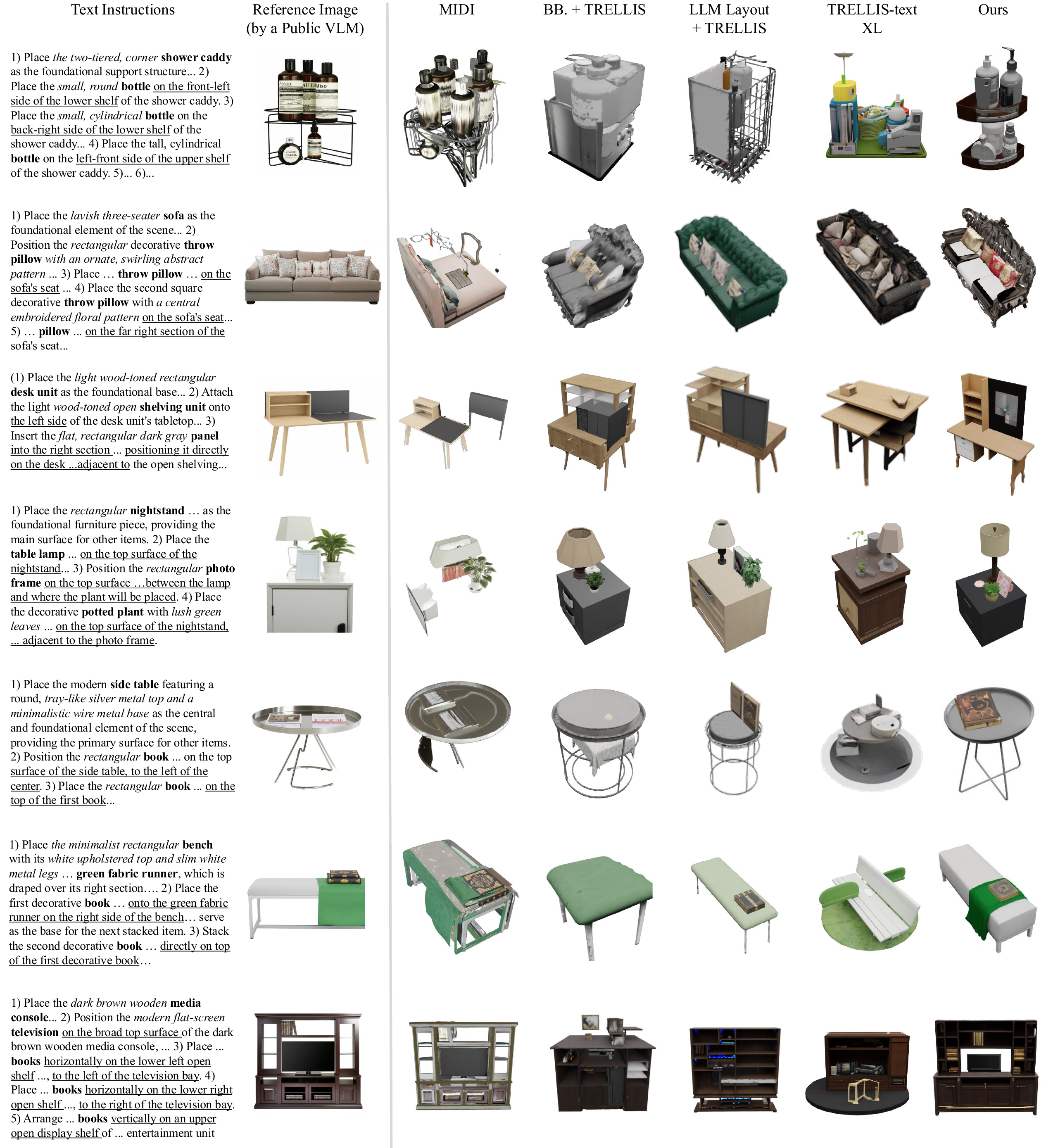}
\caption{
    \textbf{Qualitative comparison}. Visual results of our method and existing single-object generation and multi-object generation approaches are presented with corresponding input text instructions and reference images generated by a VLM. \textbf{Bold text} highlights object kinds, \textit{italic text} shape/appearance descriptions, and \underline{underline} locations. 
}
\vspace{-2mm}
\label{fig:vis_cmp}
\end{figure}

\subsection{Evaluation Settings}
We construct a held-out \rev{test set} by filtering out a subset of \evalscenesize unseen text instructions \rev{from} the training data for challenging scene composition tasks.
These scenes span diverse room types (\eg kitchen, bedroom, living room), varying spatial scales (from groups of large furniture to groups of small objects), and different object counts, ensuring comprehensive coverage of compositional complexity and spatial reasoning challenges. Our evaluation set consists of sequential text instructions captioned by a VLM, as detailed in Section~\ref{sec:data_curate}. See examples in the appendix. 

\subsection{Baselines}


\paragraph{TRELLIS-text XL.} \rev{We use the largest open-source Text-to-3D model, 2B-parameter TRELLIS~\citep{trellis2024}, as a baseline.}
To prepare the text input, we use a VLM to summarize all steps into one instruction.

\paragraph{MIDI: Multi-Instance Diffusion.} MIDI \citep{huang2025midi} is a recent image-to-scene model to generate multi-object scenes by segmenting the scene image using Grounded-SAM~\citep{ren2024grounded}, and applying a multi-instance diffusion model~\citep{huang2025midi}. To prepare the image input, we summarize all text instructions into an overall description of the scene, and use a text-to-image model to generate a scene image.

\paragraph{BB.+TRELLIS.} We  reuse the 3D bounding boxes in the test set. Then we use the TRELLIS-text XL model to generate 3D objects from object captions, and fit the generated object to the bounding box afterwards.

\paragraph{LLM-layout+TRELLIS.} In this baseline, we first use the text instructions to prompt a LLM for generating detailed object captions and predicting 3D bounding boxes. Then we use the TRELLIS-text XL model to generate actual 3D objects from those captions, rescale and position each to fit its bounding box.

%% file: sec/4_3_quali_results.tex
\subsection{Qualitative Results}

\paragraph{\ardplus sequential generation results.} Figure~\ref{fig:per_step} demonstrates how \ardplus enables creation of a compositional scene by following a sequence of text prompts. As shown in the results,  \ardplus follows text instructions at each step, and can generate object shape and appearance close to the text instructions while also placing the object based on the prescribed spatial arrangement.

\paragraph{Comparisons with baselines.} In Figure~\ref{fig:vis_cmp}, we qualitatively compare our \ardplus model with four baselines as our primary evaluation. 
\textit{TRELLIS-text XL} will hallucinate unrelated objects (\eg paper towel in row 1) or generate objects inconsistent with the text input (\eg more than expected pillows in row 2). \textit{MIDI} requires an image aligned with the text as the input. We empirically observe that when images generated from text instructions by a VLM
are used as input, MIDI may still fail to produce plausible object shapes and placements. See more details in the \rev{appendix}. For \textit{BB.+TRELLIS}, it is difficult to avoid object collision after placing individual objects, generated by TRELLIS-text XL, into the scene according to the groundtruth bounding box (\eg row 3). For \textit{LLM-layout+TRELLIS}, LLM can predict 3D bounding box of individual objects poorly for complex text input (\eg row 1 vs. 3). Even the subsequent generation of individual objects by TRELLIS-text XL are plausible, the spatial arrangement in the composed scene still significantly deviates from the text instructions. In contrast, our method successfully captures object types, shapes, appearance, and their spatial locations as specified in the text instructions, resulting in meaningful compositional 3D generation.

\paragraph{Conditioned next-object generation.}
Our model supports conditioned generation where the first 3D object is given by the user, and subsequent objects are generated based on text prompts. In our implementation, we first convert the input mesh condition into $64\times 64 \times 64$ grids, and transform the mesh using a user-input layout (specified by translation and rotation matrices) to obtain coarse-level grids. Then, we can encode the grids using our employed 3D VAE and prefill the corresponding 3D tokens with KV caching. After doing this, the subsequent generation will be conditioned on the prefilled 3D tokens of the user input. Figure~\ref{fig:gt_cond_gen} presents the results of our model for demonstrating this capability.

%% file: sec/4_4_quant_results.tex
\subsection{Quantitative Comparisons}

\begin{table}
\caption{
\rev{\textbf{Quantitative comparisons}.
We evaluate distributional quality using KID~\cite{KID2018}, P-FID, P-IS, and FD-DINOv2, and semantic alignment using CLIP~\cite{CLIP2021}. For CLIP-I, mean@20 averages scores over 20 rendered views, while best@20 reports the maximum.}
}
\centering
\rev{%
\resizebox{\columnwidth}{!}{
\begin{tabular}{lcccccccc}
\toprule
& KID ($\downarrow$)
& KDD ($\downarrow$)
& FD-DINOv2 ($\downarrow$)
& CLIP-text ($\uparrow$)
& CLIP-I mean@20 ($\uparrow$)
& CLIP-I best@20 ($\uparrow$)
& P-FID ($\downarrow$)
& P-IS ($\uparrow$) \\
\midrule
BB. + TRELLIS
& 3.036 & 158.136 & 0.407 & 19.094 & 68.774 & 77.653
& 43.706 & 4.442 \\
LLM-layout + TRELLIS
& 1.822 & 114.682 & 0.318 & 20.573 & 72.735 & 82.038
& 15.081 & 7.369 \\
TRELLIS-text XL
& 1.872 & 90.453 & 0.304 & 22.032 & 74.418 & 81.870
& 26.214 & 5.675 \\
MIDI
& 2.909 & 117.071 & 0.304 & 19.312 & 72.780 & 80.143
& 18.614 & 7.420 \\
\midrule
Ours (3D-ARD+)
& \textbf{1.791} & \textbf{55.540} & \textbf{0.280}
& \textbf{24.187} & \textbf{76.301} & \textbf{84.629}
& \textbf{8.761} & \textbf{7.454} \\
\bottomrule
\end{tabular}
}%
}
\label{tab:quanti_metrics}
\end{table}

\paragraph{Evaluation metrics.} \rev{Following prior work~\cite{trellis2024,zhang2024clay}, we choose our evaluation metrics. To evaluate generated scene quality, we evaluate 3D generations through their rendered images. We employ kernel distance metrics, including Kernel Inception Distance (KID)~\cite{KID2018} using InceptionV3 features and Kernel Distance with DINOv2 (KDD) features. We also report Fr\'echet Distance in the L2-normalized DINOv2 feature space (FD-DINOv2) of rendered images. We construct a fixed reference batch of 10K rendered images from the training set. For each of the 275 aligned test scenes, we sample 20 uniformly spaced turntable views, yielding 5.5K images per method, and select a fixed, scene-balanced subset of 5K images for KID and KDD. To measure consistency between text descriptions and generated scenes, we report CLIP-text scores averaged across all views. We use one consolidated scene description, rather than a list of instructions, as the reference text for each scene. We also provide the same description to a text-to-image model to generate one white-background reference image per scene and report CLIP-Image (mean@20) and CLIP-Image (best@20), where the latter selects the best-matching view for each scene before averaging.
For 3D-level evaluation, we report PointNet++-based FID (P-FID) and Inception Score (P-IS). The reference set consists of a fixed set of 5K training scenes, while each method generates one sample for each of the same 275 aligned test scenes. We sample 4,096 points from each reference and generated scene and normalize each point cloud by its centroid and maximum radius before extracting PointNet++ features.}

\paragraph{Results.} We present our quantitative results in Table~\ref{tab:quanti_metrics} as a secondary evaluation. Our method consistently outperforms all baseline approaches across every metric, aligning with our visual comparison. Beyond improved distribution alignment with the training data, as measured by KID, KDD, \rev{FD-DINOv2,} and P-FID, our model also achieves substantial gains in CLIP-Text, CLIP-Image, and P-IS. These metrics are computed independently of the training distribution, indicating that the improvements reflect enhanced generation quality rather than overfitting to the training data domain.

\subsection{Ablation Study}

\paragraph{Ablating refinement stage (\ard vs. \ardplus).} In Figure~\ref{fig:ablation_refinement}, we compare the results of \ard and \ardplus to illustrate the improvements introduced by the refinement stage. In the left example (the ornament case), we show \ardplus generates more fine-grained objects with the designed refinement step. The right example (the sofa case) shows that generating appearance directly on coarse geometry is likely to cause artifacts in the resulting texture. In addition, the shape of the pillow after the refinement stage also has smoother geometry, showing the effectiveness of our approach.

\paragraph{Ablating object-level data and patch size.} We conduct an ablation study on two key factors that affect generation quality: (1) whether single-object data is included in the training set, and (2) the patch size used for 3D generation tokens in the transformer. All models in this ablation study use the same hyperparameters and model size as our final model described in \rev{Section~\ref{sec:impl_details}}. Table~\ref{tab:ablation_ps_objdata} summarizes the results.
The results suggest two main observations. First, even without object-level training data, the model can produce reasonable spatial layouts guided by text instructions, as indicated by competitive CLIP-Text scores. However, the P-FID metric reveals that the generated shapes are of low quality. Second, our experiments show that using a patch size of 2 for generation tokens (the exact setting used for 3D understanding tokens) introduces ``patching artifacts'', leading to severe holes in the decoded 3DGS results. Therefore, we adopt a patch size of 1 for generation tokens, which significantly improves the final generation quality.

\begin{table}[t]
\caption{\textbf{Ablation study} of object-level training data and patch size, using CLIP-text and P-FID scores. \rev{All variants are evaluated on the same 50 held-out test scenes, using random rendered views for CLIP-text and 5K test examples as the P-FID reference.}}
\centering
\resizebox{0.6\columnwidth}{!}{
\begin{tabular}{lcc}
\toprule
Method & CLIP-text ($\uparrow$) & P-FID ($\downarrow$) \\
\midrule
w/o object-level training data & 24.772 & 6.238 \\
patch size for gen. tokens = 2 & 24.647 & 4.521 \\ 
\midrule 
Final Model & \textbf{27.447} & \textbf{2.837} \\ 
\bottomrule
\end{tabular}
}
\label{tab:ablation_ps_objdata}
\vspace{-0.15in}
\end{table}

%% file: sec/5_conclusion.tex
\section{Conclusion}
We present a novel Autoregressive 3D Diffusion model \ardplus for compositional \rev{text-to-scene generation by jointly modeling} layout and shape. A large-scale indoor scene dataset with paired text instructions is curated to train the \ardplus model. On a challenging evaluation set, the proposed \ardplus model significantly outperforms other competing methods in both visual comparisons and quantitative evaluations.
Our model can serve as a base model for a series of future post-training work, such as physics-informed supervised fine-tuning for generating more realistic scenes.

%% file: sec/main_supp.tex



\clearpage

\beginappendix
\beginsupplement


\input{sec/supp_qualitative_res}
\input{sec/supp_ablative_study}
\input{sec/supp_data_curation}
\input{sec/supp_impl_details}
\input{sec/supple_eval_set}
\input{sec/supp_limitation}

\begin{figure}[t]
    \centering
    \includegraphics[width=\linewidth]{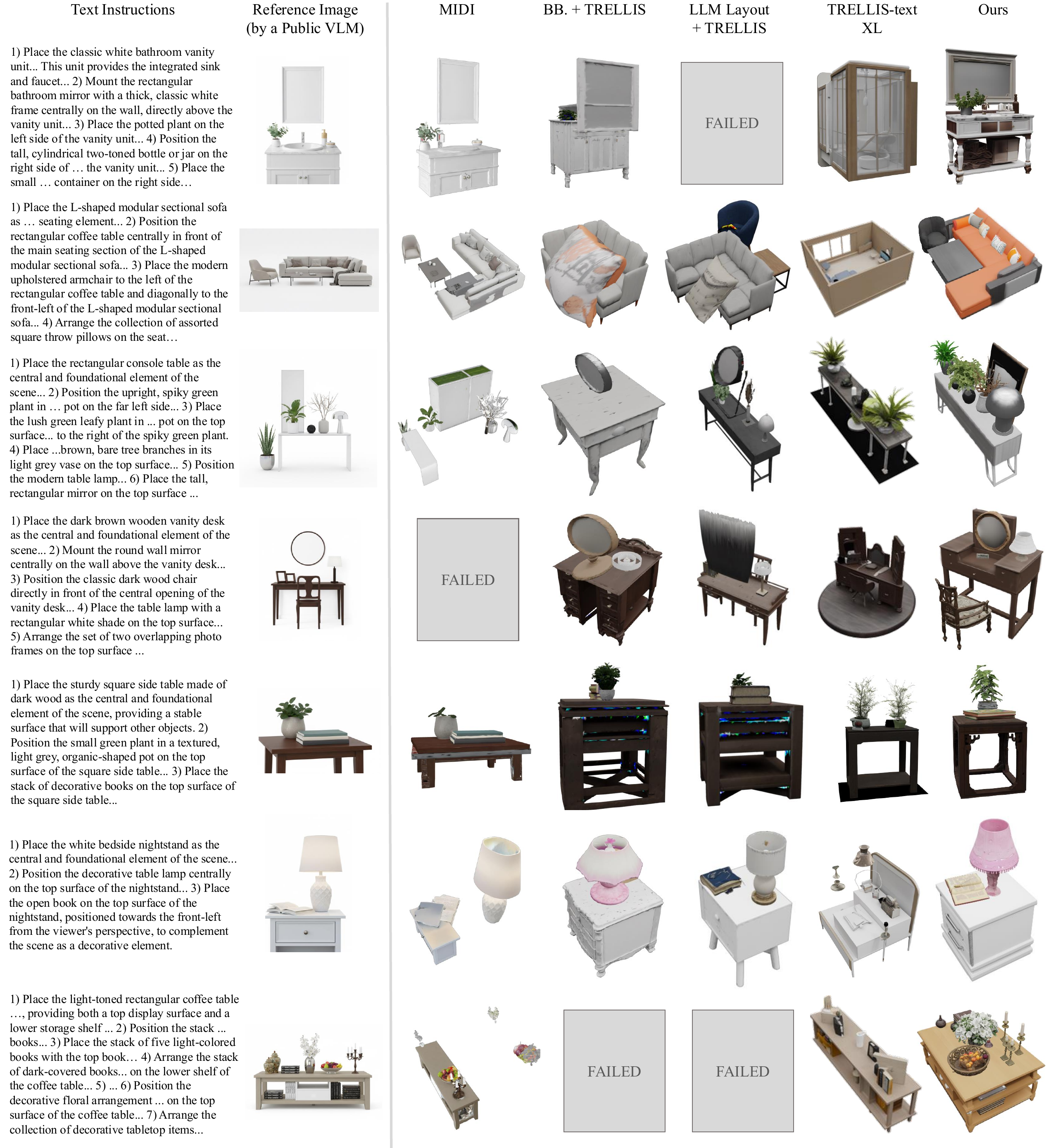}
    \caption{\textbf{More qualitative comparison results.} We present additional visual results comparing our method with baseline approaches. Grey boxes labeled "FAILED" indicate that the corresponding baseline methods are unable to produce any output.}
    \label{fig:more-quali}
\end{figure}

%% file: sec/supp_qualitative_res.tex
\section{More Results}








\subsection{Diversity of Generation}

\begin{figure}[t]
    \centering
    \includegraphics[width=\linewidth]{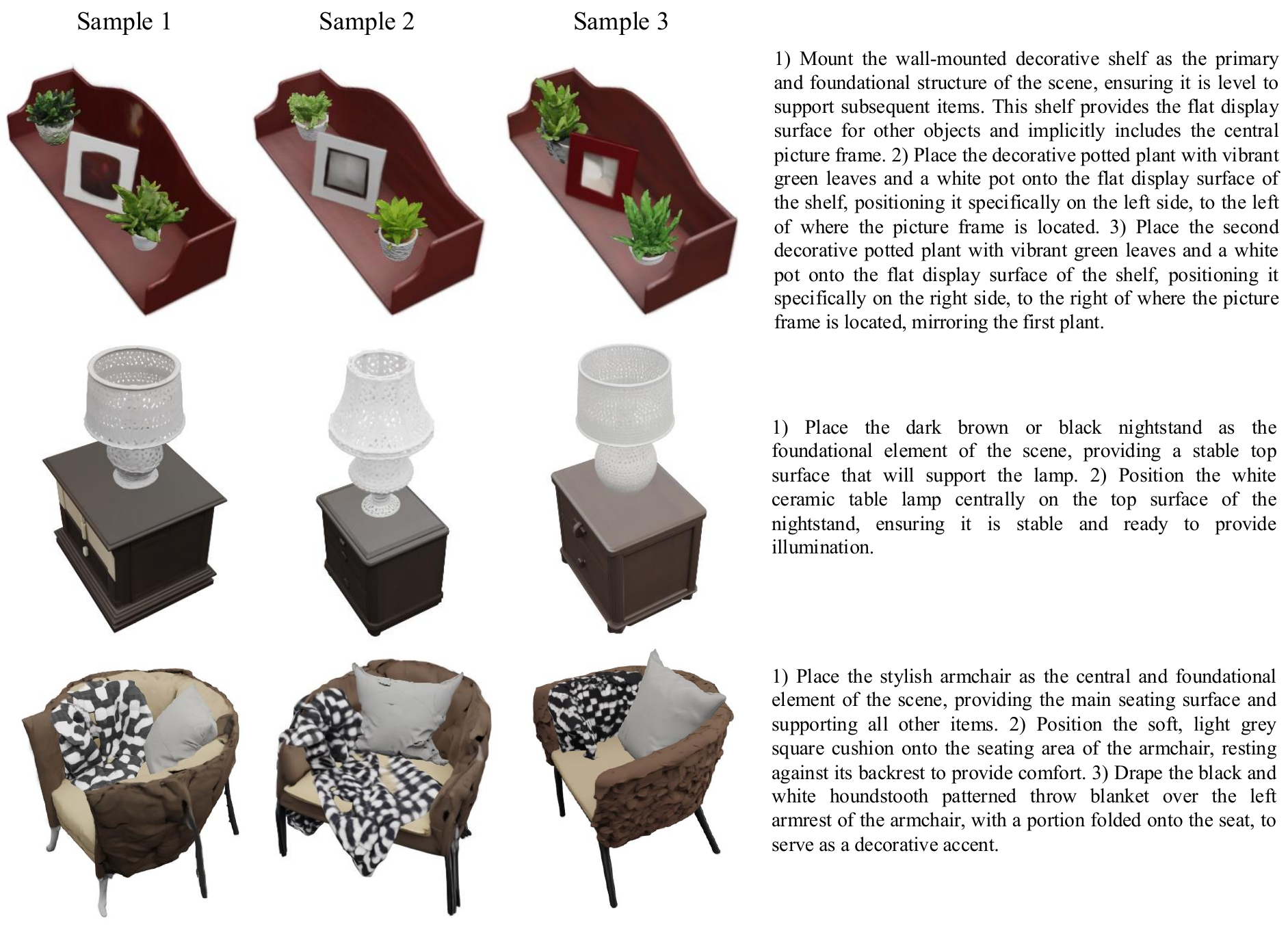}
    \caption{Diversity of generation results. Three samples are generated for
    each sequence of text instructions, illustrating the variations of output produced
    by our method. The first-row results are generated via conditioning on the
    ground-truth geometry of the base object. Results in the second and third rows
    are generated from scratch. }
    \vspace{-2mm}
    \label{fig:diversity}
\end{figure}

To demonstrate the generative diversity of our model, we provide multiple generations with different random seeds from identical text instructions. Figure~\ref{fig:diversity} shows that given the same prompt, our model produces varied yet semantically consistent results.

\subsection{Additional Qualitative Results}

In addition to the comparisons provided in the main text, Figure~\ref{fig:more-quali} presents further generation results produced by our method and various baseline approaches. In the presented cases, our method demonstrates superior visual quality in synthesizing compositional scenes that follow the provided text instructions, outperforming the baseline approaches. Furthermore, several baseline methods, including MIDI, BB. + TRELLIS, and LLM Layout + TRELLIS, fail to generate output for certain instructions (either empty output or crashes when querying models), revealing the instability of the baseline systems.

%% file: sec/supp_data_curation.tex
\begin{figure}[t]
\centering
\begin{subfigure}[b]{0.24\columnwidth}
    \centering
    \includegraphics[width=\textwidth]{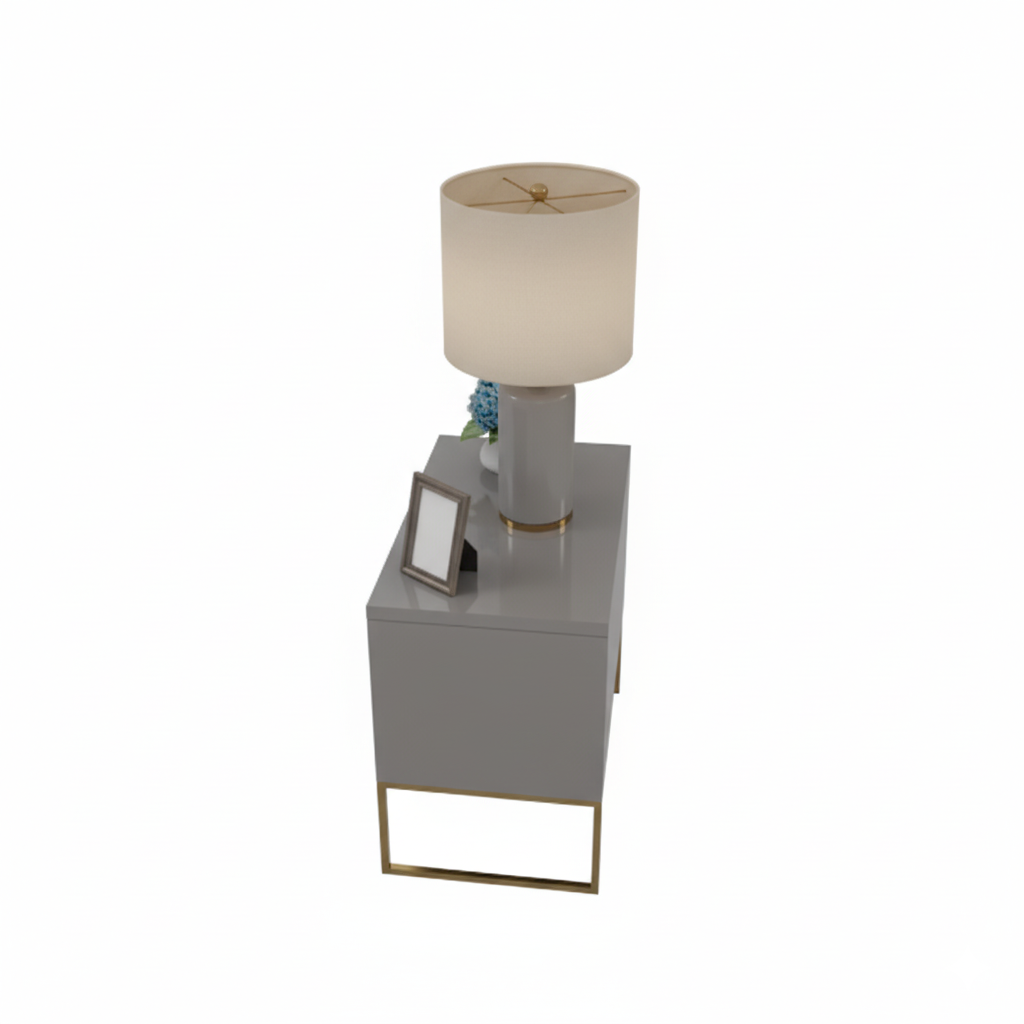}
    \caption{View 0}
    \label{fig:view0}
\end{subfigure}
\hfill
\begin{subfigure}[b]{0.24\columnwidth}
    \centering
    \includegraphics[width=\textwidth]{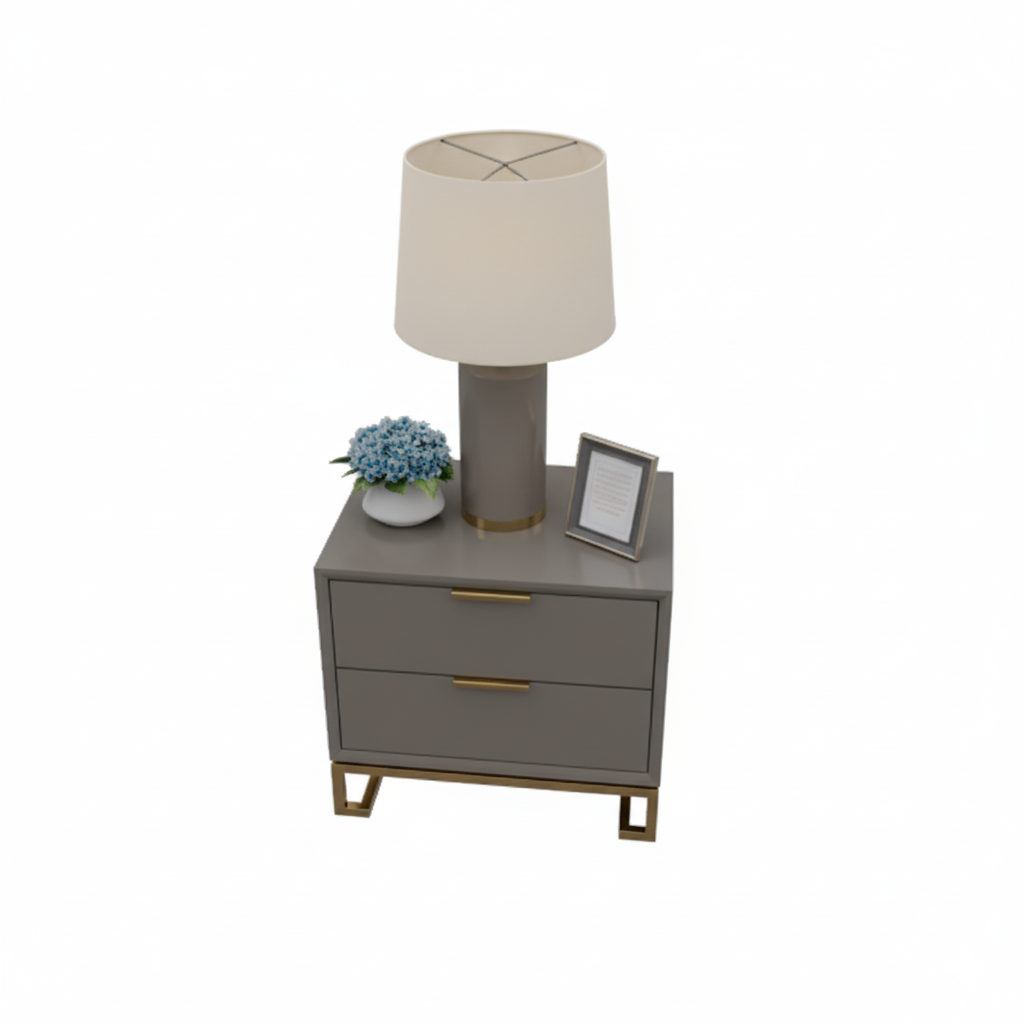}
    \caption{View 1 (Selected)}
    \label{fig:view1}
\end{subfigure}
\hfill
\begin{subfigure}[b]{0.24\columnwidth}
    \centering
    \includegraphics[width=\textwidth]{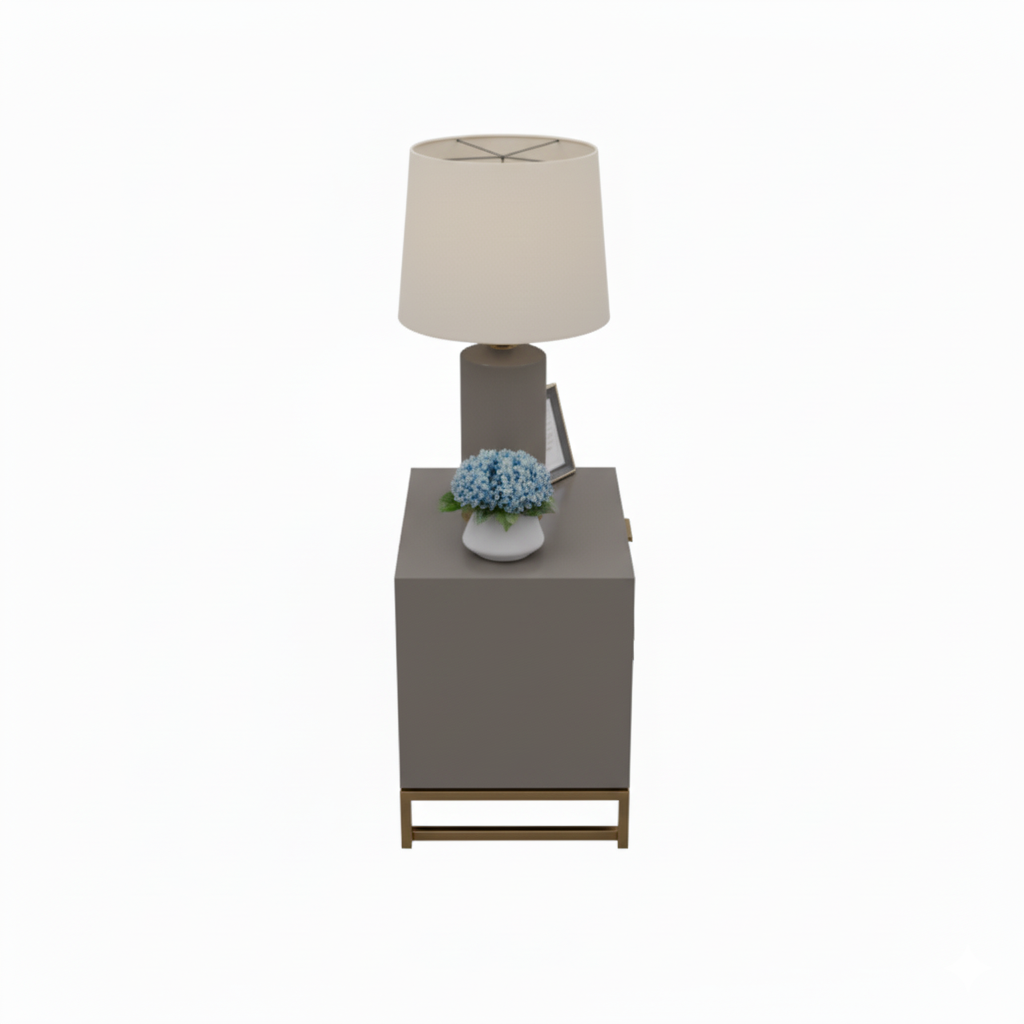}
    \caption{View 2}
    \label{fig:view2}
\end{subfigure}
\hfill
\begin{subfigure}[b]{0.24\columnwidth}
    \centering
    \includegraphics[width=\textwidth]{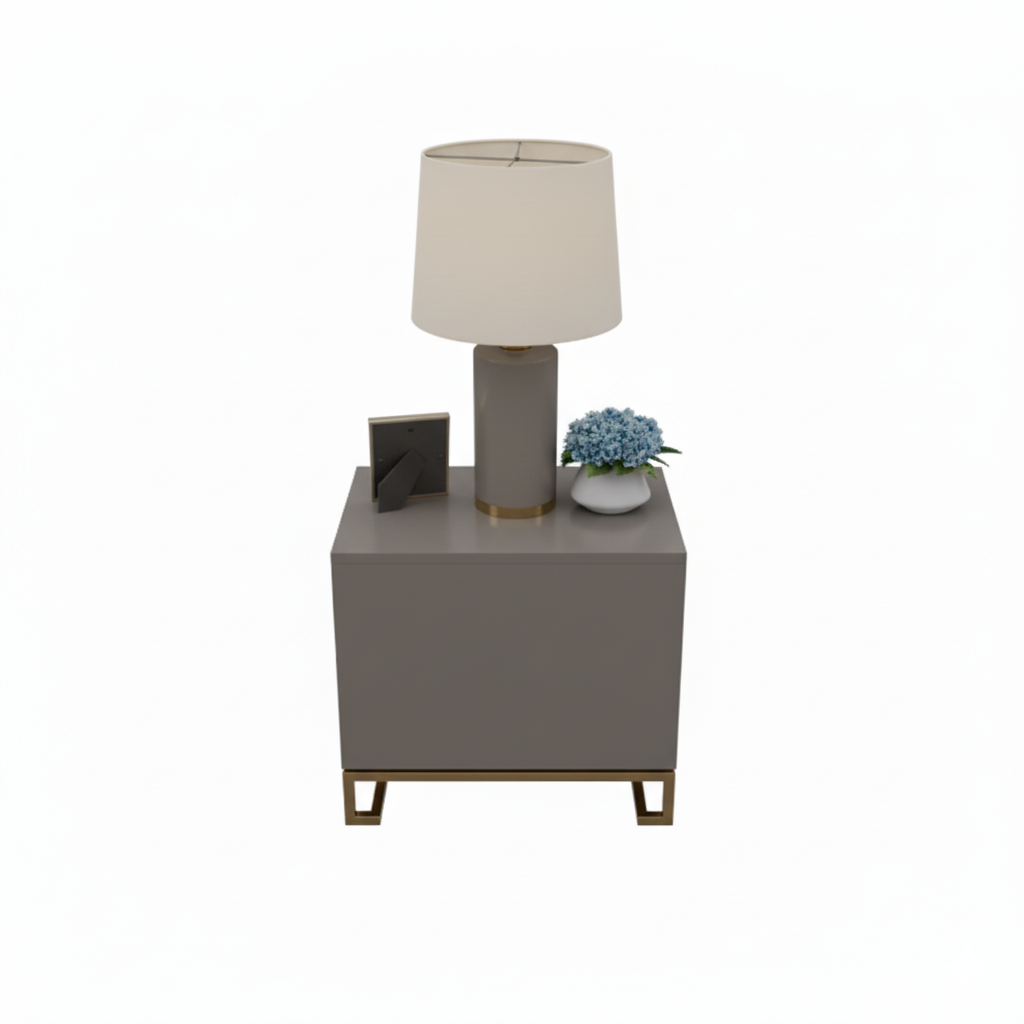}
    \caption{View 3}
    \label{fig:view3}
\end{subfigure}

\caption{View selection example. Given four different viewpoint renderings of the same scene, our method selects View 1 as the canonical front view.}
\label{fig:view_selection}
\end{figure}

\begin{figure}[t]
\centering
\begin{subfigure}[b]{0.23\columnwidth}
    \centering
    \includegraphics[width=\textwidth, height=0.8\columnwidth, keepaspectratio]{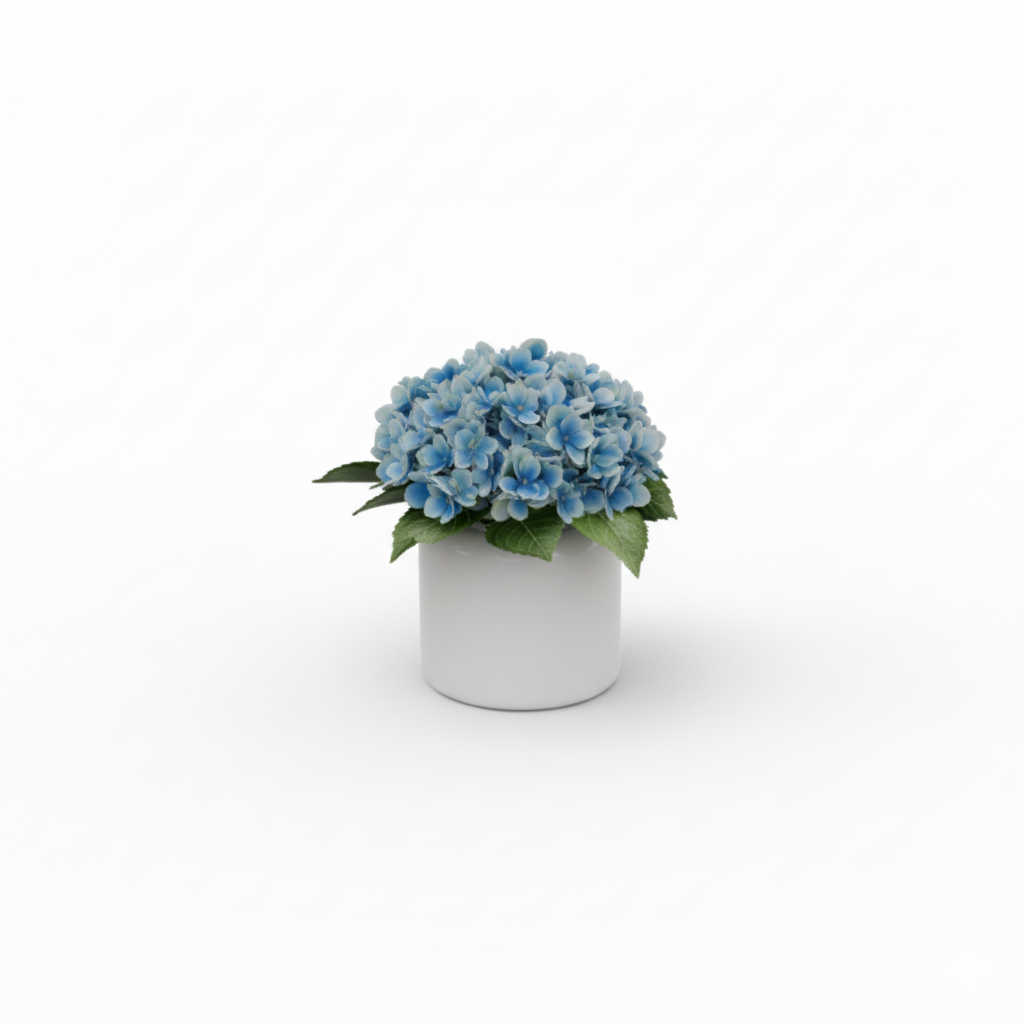}
    \caption{Part 0}
    \label{fig:norm0}
\end{subfigure}
\hfill
\begin{subfigure}[b]{0.23\columnwidth}
    \centering
    \includegraphics[width=\textwidth, height=0.8\columnwidth, keepaspectratio]{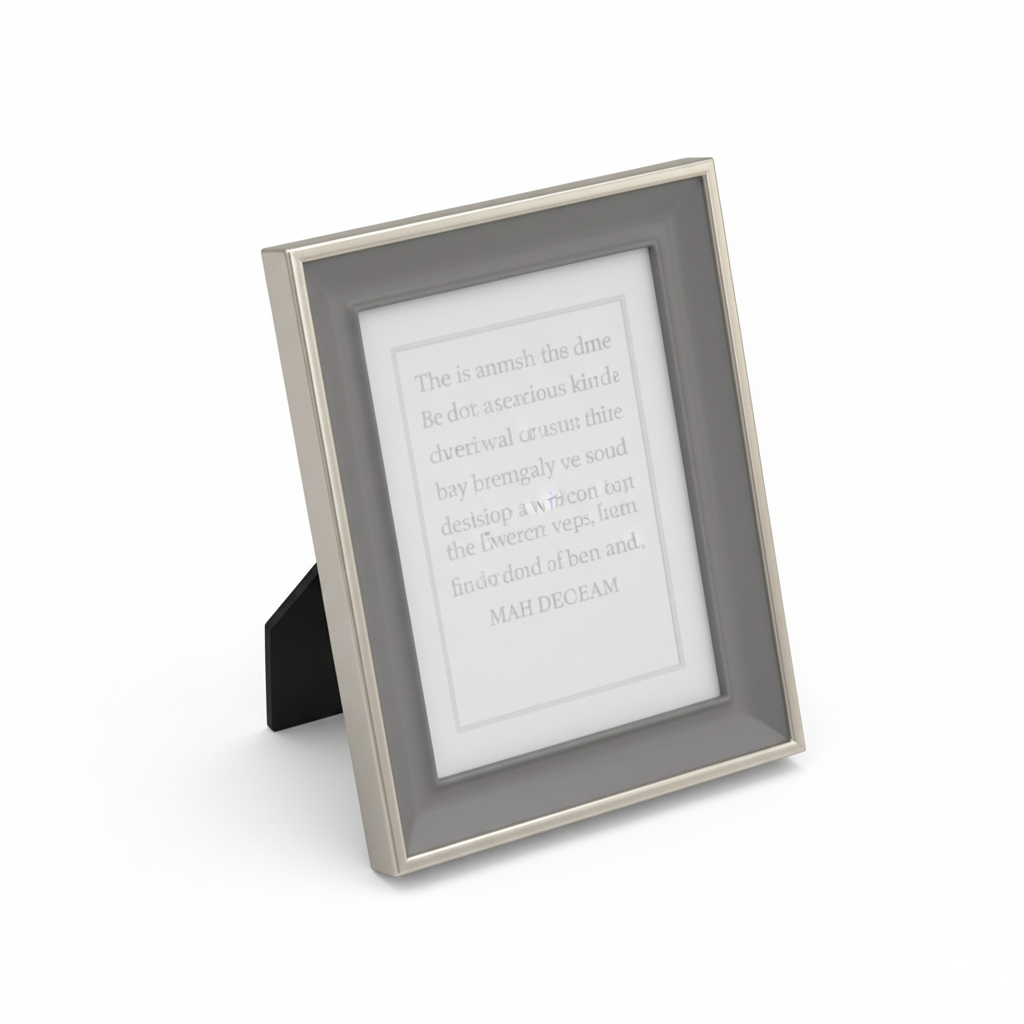}
    \caption{Part 1}
    \label{fig:norm1}
\end{subfigure}
\hfill
\begin{subfigure}[b]{0.23\columnwidth}
    \centering
    \includegraphics[width=\textwidth, height=0.8\columnwidth, keepaspectratio]{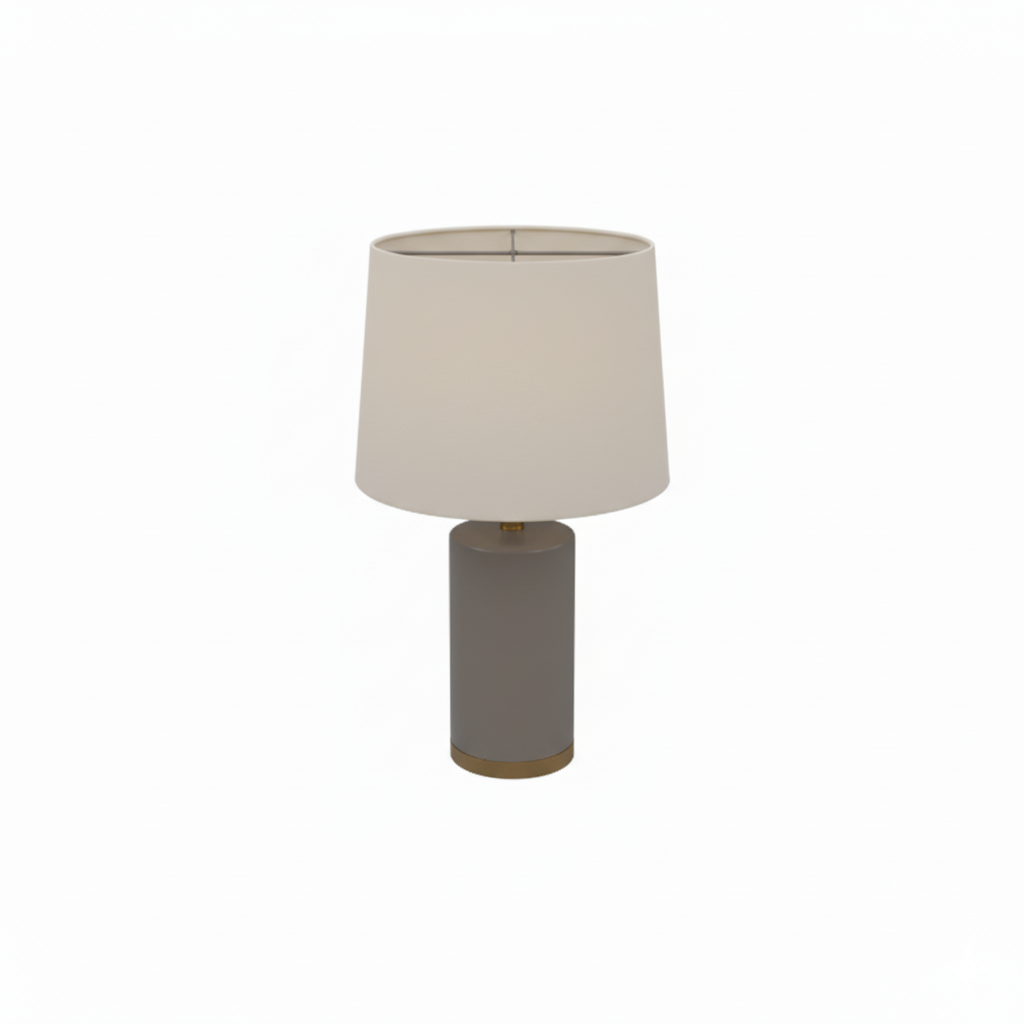}
    \caption{Part 2}
    \label{fig:norm2}
\end{subfigure}
\hfill
\begin{subfigure}[b]{0.23\columnwidth}
    \centering
    \includegraphics[width=\textwidth, height=0.8\columnwidth, keepaspectratio]{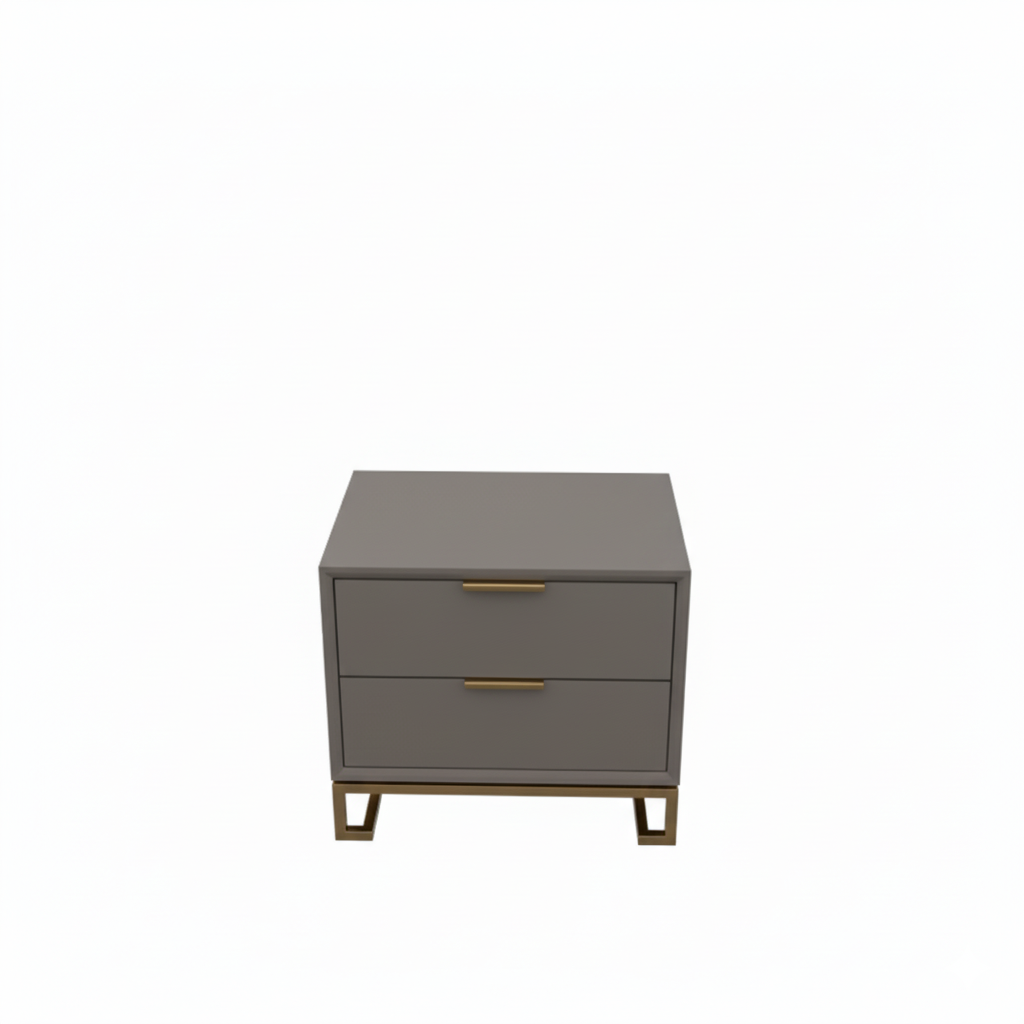}
    \caption{Part 3}
    \label{fig:norm3}
\end{subfigure}
\hfill
\caption{Normalized input views for object captioning. Each part is extracted and normalized from the selected front view to provide clear, focused views for detailed caption generation.}
\label{fig:norm_views}
\end{figure}

\section{Data Curation Details}


Our data curation pipeline leverages a VLM for all stages of processing, including view selection, object captioning, and assembly plan generation. Below we detail the specific prompts and procedures used at each stage.

Throughout this section, we use a nightstand scene as a running example to illustrate each stage of our data curation pipeline. This scene consists of four objects: Part 0 is a blue hydrangea potted plant, Part 1 is a rectangular picture frame, Part 2 is a modern table lamp, and Part 3 is a grey nightstand with a gold base (the foundational furniture piece).

\subsection{View Selection}

To identify the canonical front view from multi-view renderings, we provide the following system instructions along with 4 different viewpoint images. Figure~\ref{fig:view_selection} shows an example where our method selects the most informative viewing angle from four candidate views. The complete prompt is shown in Figure~\ref{fig:view_selection_prompt}.

\begin{figure}[t]
\begin{tcolorbox}[colback=gray!10, colframe=gray!50, boxrule=0.5pt, title=View Selection Prompt]
\small
\textbf{System Instruction:} You are an indoor scene analyzer. Given 4 different viewpoint images of the same indoor scene or object group, analyze them carefully. Your task is to identify the front view. The 'front view' is the most informative viewing angle, showing almost all parts of the main object or scene with minimal occlusion.

\textbf{Domain-specific criteria:}
\begin{itemize}
    \item If the object is a wardrobe or cabinet, the side with doors or visible contents is the front view (note that you should distinguish what is the back side of the wardrobe, and filter this).
    \item If it is a bed, the side with pillows and bedside tables should be at the top, farthest away from the camera, receding into the background.
    \item If it is a desk or table, the side where the user can most easily interact directly is considered the front view.
\end{itemize}

In this task, you must choose the front view from a slightly top-down perspective (not a perfectly level/straight-on view). The selection should prioritize the angle that would be considered the canonical or primary viewing direction for recognizing or interacting with the scene or object.

\textbf{Analysis format:}
\begin{enumerate}
    \item Whether they include wardrobe, cabinet: True/False
    \begin{itemize}
        \item If 1 is True, whether the image includes the doors (sometimes the backside looks like doors, should be careful!), whether it includes contents and what contents are they, whether the user is easy to get access to everything of the cabinet or on the table
    \end{itemize}
    \item Whether they include beds: True/False
    \begin{itemize}
        \item If 2 is True, whether there are bedside tables
        \item If 2 is True, whether the side with pillows and bedside tables is at the far end: True/False
    \end{itemize}
    \item Small objects that you can see from the image
    \item If multiple images meet the above requirements, compare and describe them and find the one that meets most
\end{enumerate}

\textbf{User Prompt:} Finally the response should end in this format: ``the front view is x'', where x should be 0, 1, 2, 3.

Here are the four views: [000.png] [001.png] [002.png] [003.png].
\end{tcolorbox}
\caption{View selection prompt used to identify the canonical front view from multi-view renderings.}
\label{fig:view_selection_prompt}
\end{figure}

\begin{figure}[t]
\centering
\includegraphics[width=0.8\columnwidth]{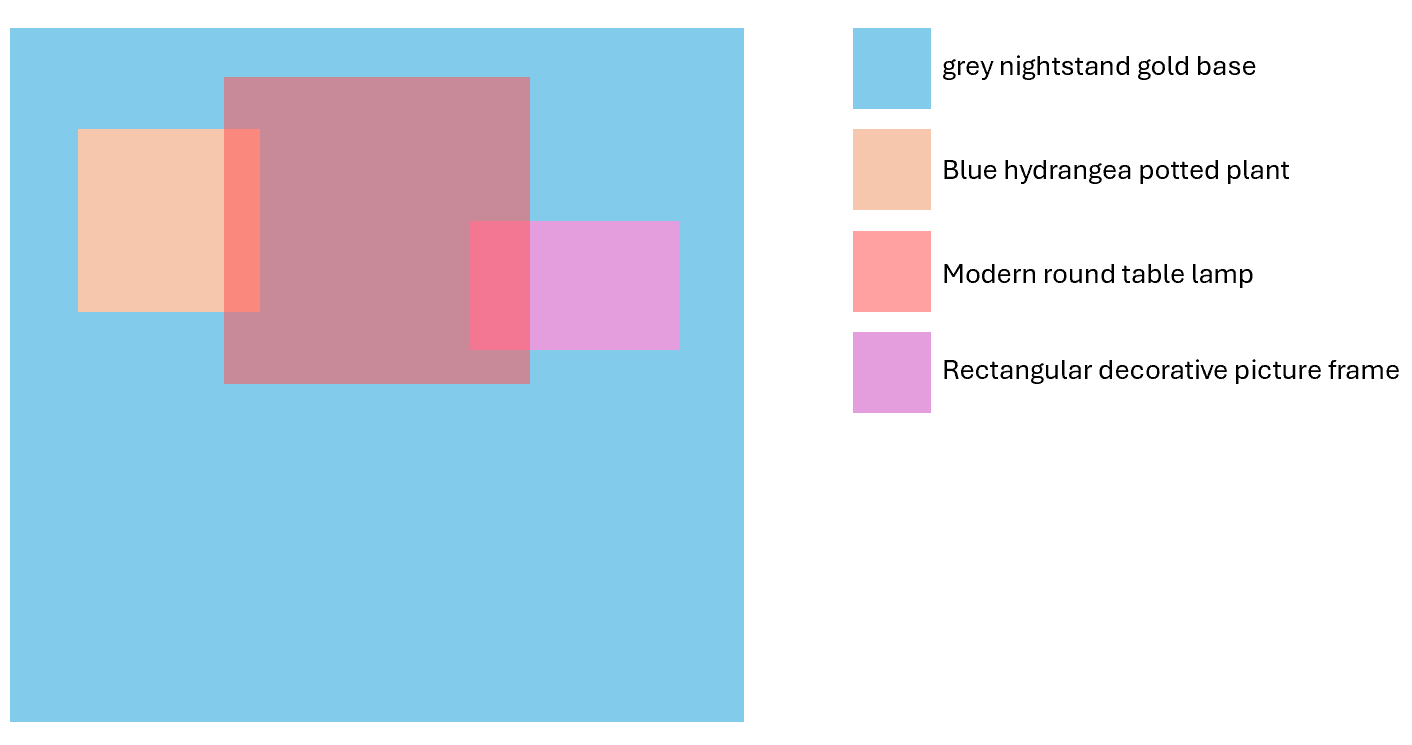}
\caption{Bounding box visualization for spatial relationship analysis. The top-down view shows the X-Z plane projection with all objects represented by their bounding box footprints. The legend is labeled with short captions.}
\label{fig:bbox_vis}
\end{figure}

\begin{figure*}[t]
\centering
\begin{subfigure}[b]{0.24\textwidth}
    \centering
    \includegraphics[width=\textwidth]{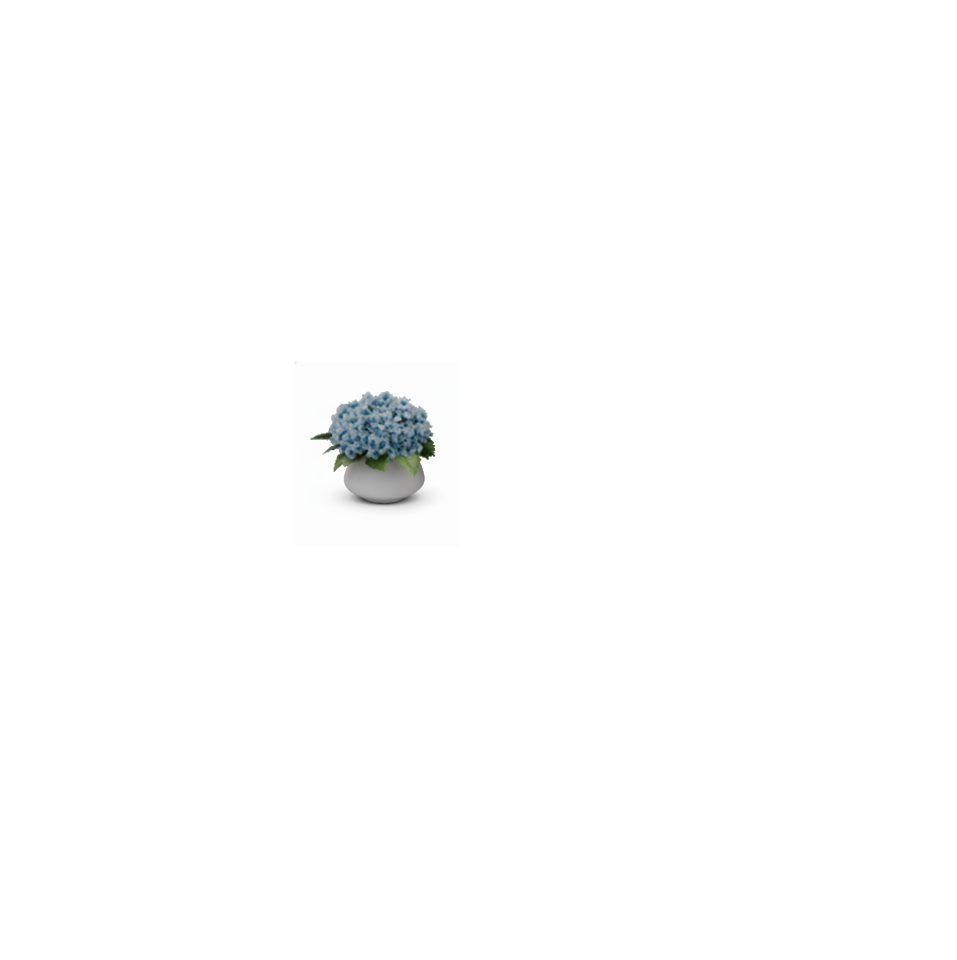}
    \caption{Part 0}
    \label{fig:ind0}
\end{subfigure}
\hfill
\begin{subfigure}[b]{0.24\textwidth}
    \centering
    \includegraphics[width=\textwidth]{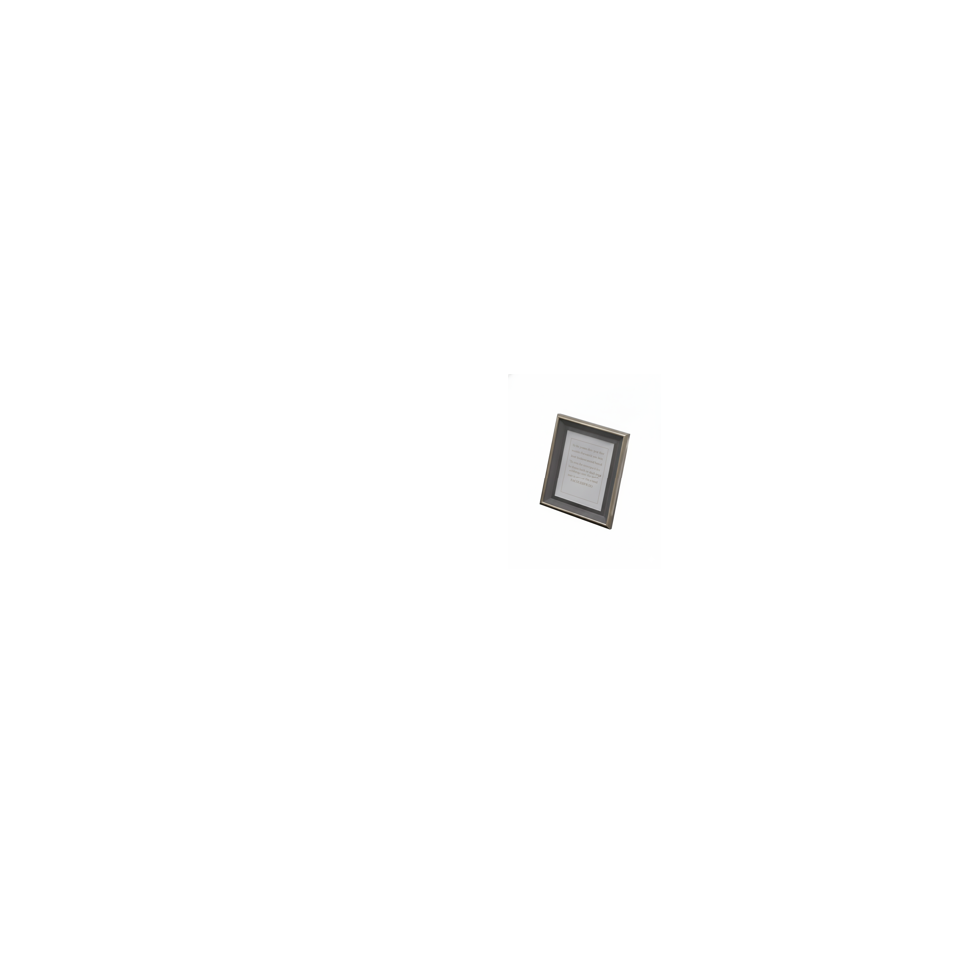}
    \caption{Part 1}
    \label{fig:ind1}
\end{subfigure}
\hfill
\begin{subfigure}[b]{0.24\textwidth}
    \centering
    \includegraphics[width=\textwidth]{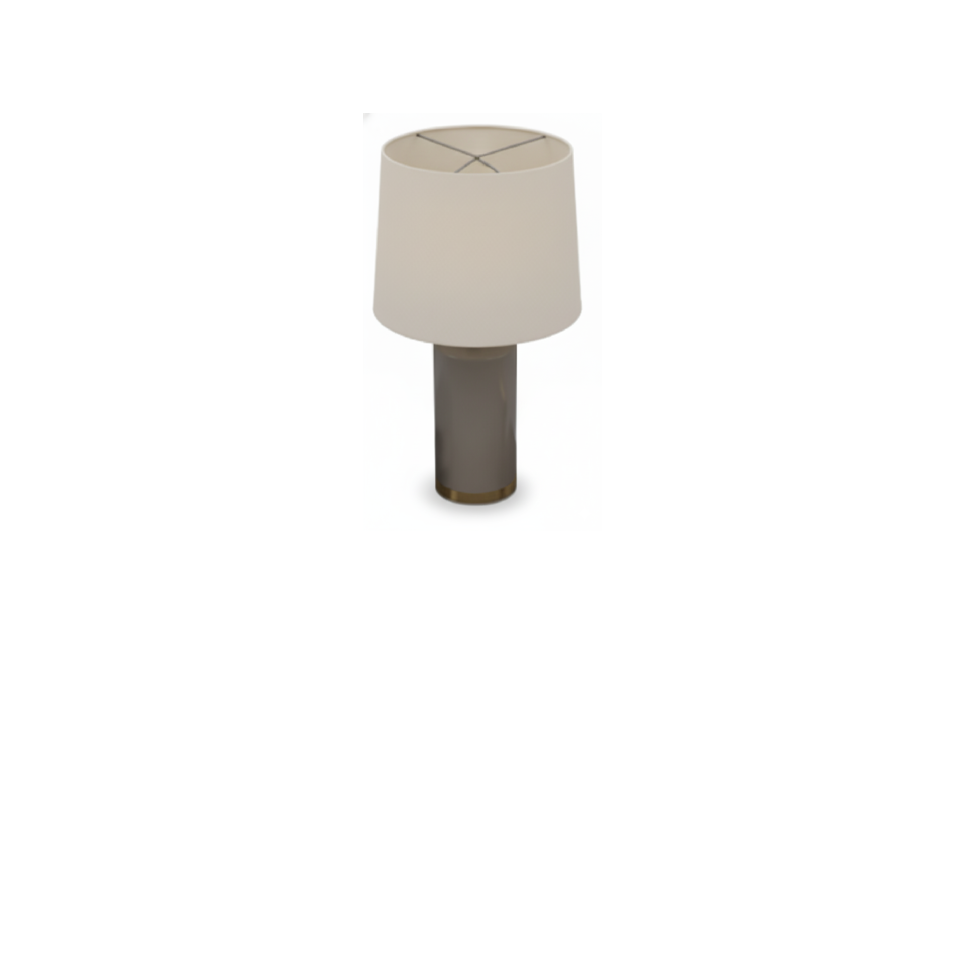}
    \caption{Part 2}
    \label{fig:ind2}
\end{subfigure}
\hfill
\begin{subfigure}[b]{0.24\textwidth}
    \centering
    \includegraphics[width=\textwidth]{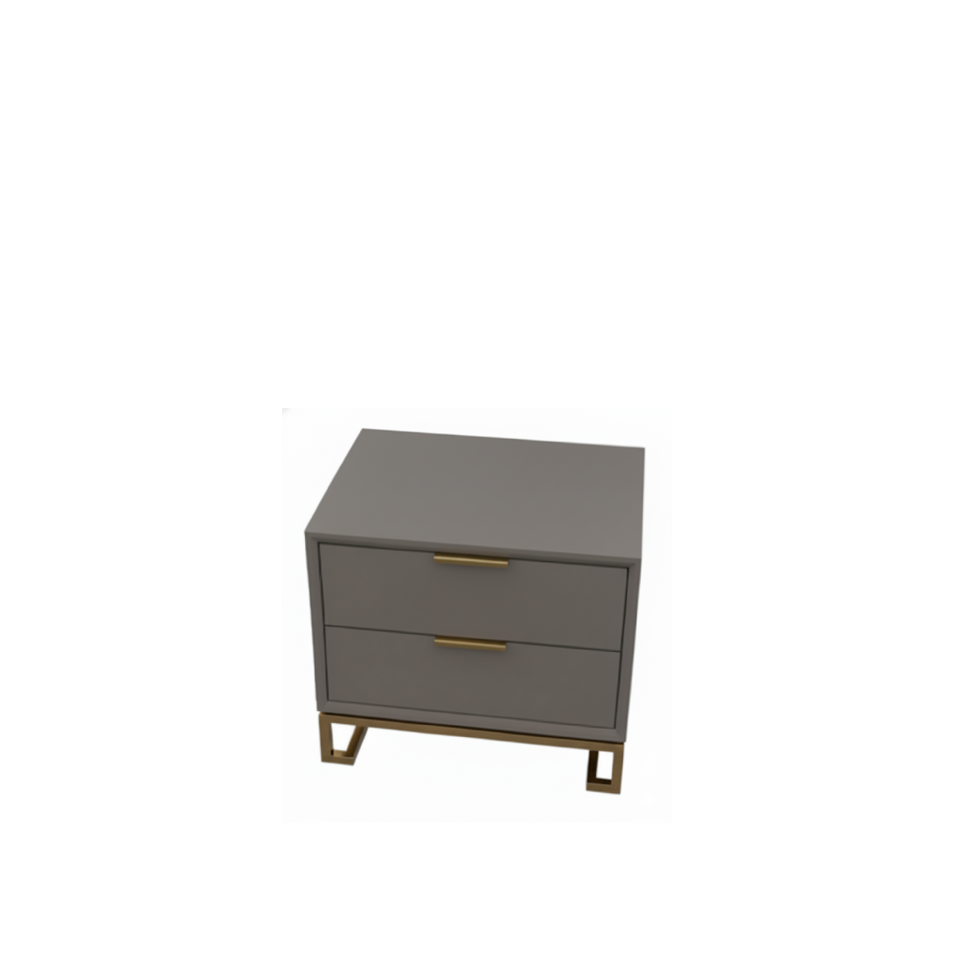}
    \caption{Part 3}
    \label{fig:ind3}
\end{subfigure}

\vspace{0.2cm}

\begin{subfigure}[b]{0.24\textwidth}
    \centering
    \includegraphics[width=\textwidth]{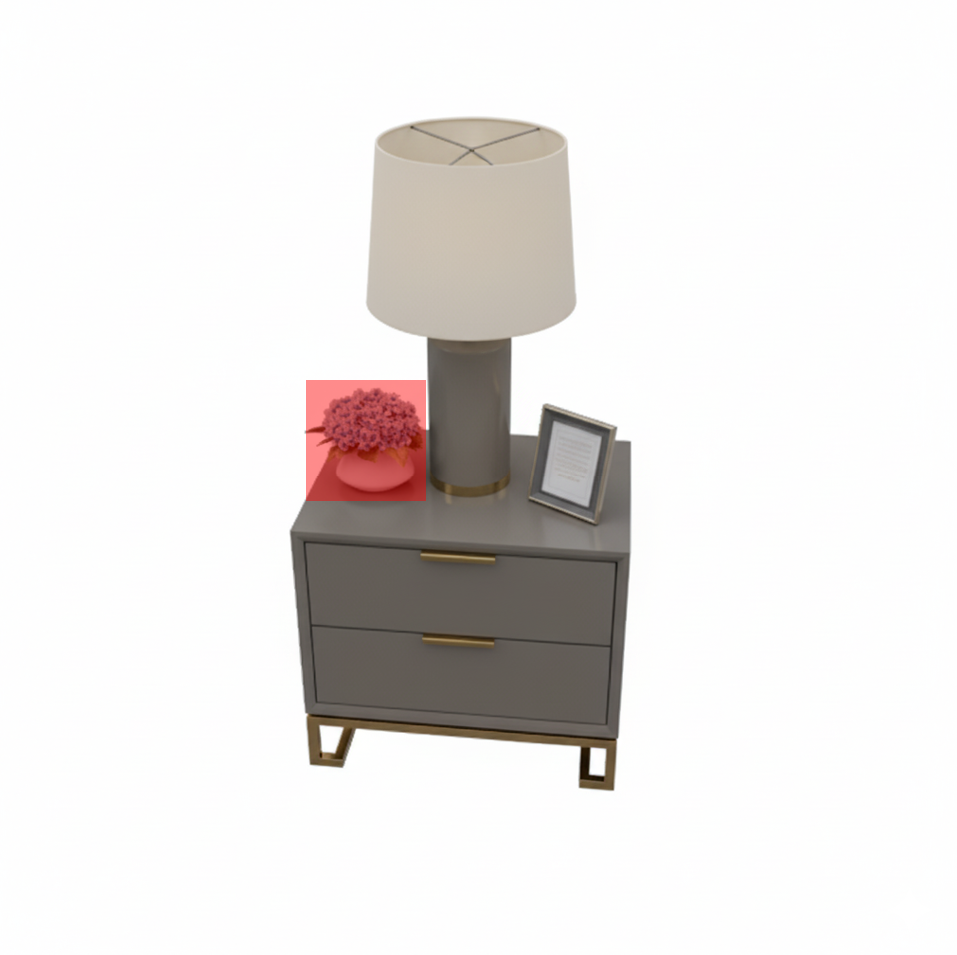}
    \caption{Part 0 (mask)}
    \label{fig:mask0}
\end{subfigure}
\hfill
\begin{subfigure}[b]{0.24\textwidth}
    \centering
    \includegraphics[width=\textwidth]{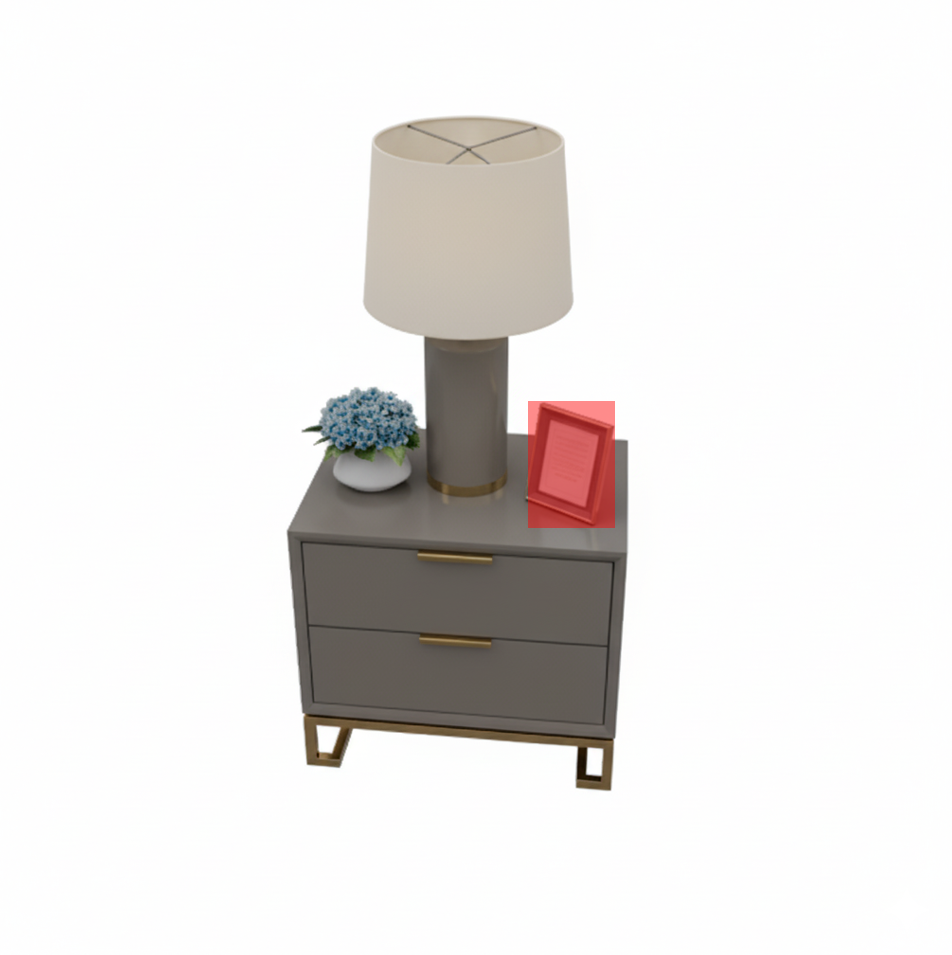}
    \caption{Part 1 (mask)}
    \label{fig:mask1}
\end{subfigure}
\hfill
\begin{subfigure}[b]{0.24\textwidth}
    \centering
    \includegraphics[width=\textwidth]{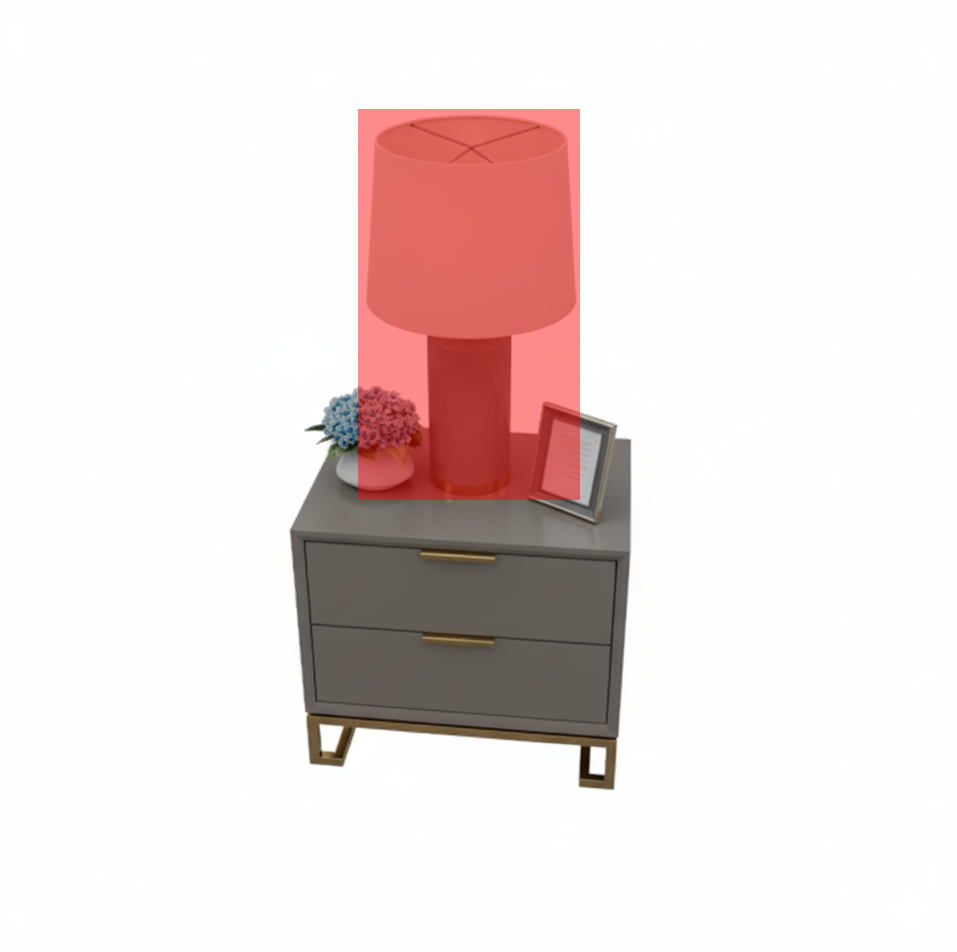}
    \caption{Part 2 (mask)}
    \label{fig:mask2}
\end{subfigure}
\hfill
\begin{subfigure}[b]{0.24\textwidth}
    \centering
    \includegraphics[width=\textwidth]{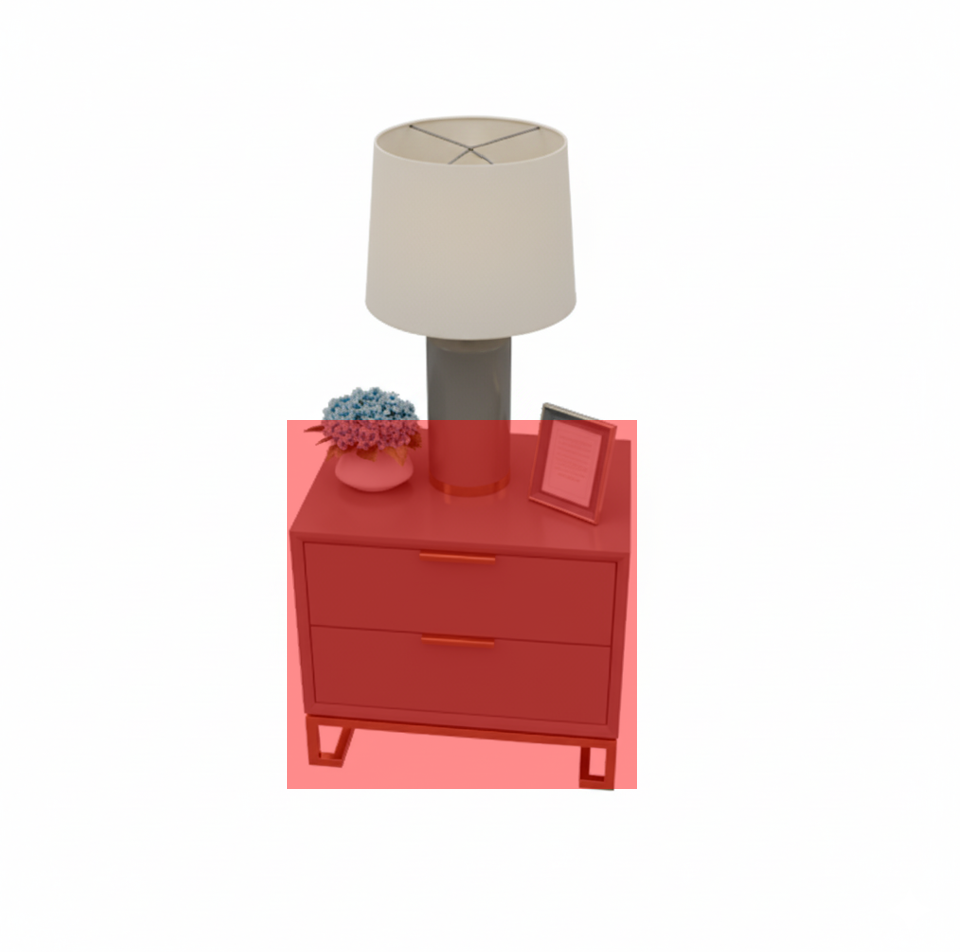}
    \caption{Part 3 (mask)}
    \label{fig:mask3}
\end{subfigure}

\caption{Input visualization for assembly plan generation. Top row: Individual parts in its original location. Bottom row: masks highlighting each part in red for precise object identification. These visualizations, combined with bounding box data (Figure~\ref{fig:bbox_vis}), enable the model to reason about spatial relationships and generate physically grounded assembly instructions.}
\label{fig:assembly_inputs}
\end{figure*}

\subsection{Object Captioning}

For each object in the scene, we generate multi-granularity (one in around 20 words, another in around 5 words) textual descriptions. The model receives two types of images: (1) the entire scene view (all\_together.png, using the selected front view from Figure~\ref{fig:view_selection}) showing the complete furniture group, and (2) normalized views of each part (norm\_part\_x.png) as shown in Figure~\ref{fig:norm_views}. The complete prompt is shown in Figure~\ref{fig:captioning_prompt}.

\begin{figure}[t]
\begin{tcolorbox}[colback=gray!10, colframe=gray!50, boxrule=0.5pt, title=Object Captioning Prompt]
\tiny
\textbf{System Instruction:} You are an indoor scene analyzer specializing in furniture and object analysis with focus on assembly planning. You will receive two types of images for each part:
\begin{enumerate}
    \item Entire scene view with all objects together (showing the complete furniture group/scene)
    \item Normalized/magnified views of each part (zoomed-in view focusing on the specific part)
\end{enumerate}

Your goal is to generate a concise descriptive caption for each part, considering its role in the assembly process.

\textbf{IMPORTANT:} If multiple parts are the same asset but in different location/rotation, please use exactly the same description.

\textbf{IMPORTANT:} Consider assembly steps when captioning. If a piece of furniture is supporting other objects, describe those supporting structures and capabilities.

\textbf{For each part, describe:}
\begin{itemize}
    \item Object type/component (e.g., chair, table leg, drawer, door, etc.)
    \item Material, color, and key visual features
    \item Shape and style characteristics
    \item Supporting structures and surfaces (if the object supports other items)
\end{itemize}

\textbf{Assembly-focused caption guidelines:}
\begin{itemize}
    \item For tables, desks, shelves: Describe the surface area, load-bearing capacity indicators
    \item For chairs, stools: Describe seating surface, backrest, armrests
    \item For storage furniture: Describe compartments, doors, drawers, and internal organization
    \item For supporting structures: Describe legs, frames, bases, and their stability characteristics
    \item There should be another \texttt{short\_caption}, which is a condensed version of the caption in approximately 5 words, capturing the essential object type and key feature (e.g., `wooden dining chair', `round glass table', `metal desk lamp')
\end{itemize}

\textbf{Format your response as a JSON object:}

\texttt{\{"captions": [\{"part\_idx": 0, "caption": "concise assembly-aware description", "short\_caption": "5-word summary"\}, \{"part\_idx": 1, "caption": "...", "short\_caption": "..."\}, ...\}]}

\textbf{CRITICAL REQUIREMENTS for captions:}
\begin{itemize}
    \item Each item must have exactly three keys: `part\_idx', `caption', `short\_caption'
    \item Each part\_idx from 0 to N-1 (where N is total parts) must appear exactly once
    \item The part\_idx=x description must correspond STRICTLY to the x-th normalized image shown in the input sequence
\end{itemize}

Keep captions concise but descriptive. Focus on visual appearance, object identification, and assembly functionality.

\textbf{Here is an example of the expected format:}

\begin{lstlisting}[language=json, basicstyle=\tiny\ttfamily]
{
  "captions": [
    {
      "part_idx": 0,
      "caption": "A round side table with a dark wood top and a black metal splayed leg frame, serving as the base for the scene.",
      "short_caption": "round side table"
    },
    {
      "part_idx": 1,
      "caption": "A decorative group consisting of two abstract human-like figurines, one black and one white, seated next to a black lantern with a white candle.",
      "short_caption": "decorative figurine group"
    },
    {
      "part_idx": 2,
      "caption": "A black table lamp featuring a pleated shade and an angular, faceted base, providing illumination.",
      "short_caption": "black table lamp"
    }
  ]
}
\end{lstlisting}

\textbf{User Prompt:} Generate captions for each part and provide the response in JSON format.

Images: [all\_together.png], [norm\_part\_0.png], [norm\_part\_1.png], ...
\end{tcolorbox}
\caption{Object captioning prompt used to generate multi-granularity textual descriptions for each object in the scene.}
\label{fig:captioning_prompt}
\end{figure}

\begin{figure}[t]
\centering
\begin{lstlisting}[language=json, basicstyle=\tiny\ttfamily]
{
  "captions": [
    {
      "part_idx": 0,
      "caption": "A decorative potted plant featuring a dense arrangement of light blue hydrangea flowers and green leaves in a smooth, round white vase, designed to be placed on a flat surface.",
      "short_caption": "Blue hydrangea potted plant"
    },
    {
      "part_idx": 1,
      "caption": "A rectangular picture frame with an elegant, molded light beige finish and a white image, featuring a kickstand for freestanding tabletop display.",
      "short_caption": "Rectangular decorative picture frame"
    },
    {
      "part_idx": 2,
      "caption": "A modern table lamp featuring a tapered round light-colored fabric shade with dark trim, mounted on a sturdy, angular dark brown or black square base, providing illumination.",
      "short_caption": "Modern round table lamp"
    },
    {
      "part_idx": 3,
      "caption": "A sleek, grey nightstand or cabinet featuring a flat top surface for display, two drawers with a gold-toned rectangular pull handle, and supported by a striking gold-colored metal base, providing stability and storage.",
      "short_caption": "grey nightstand gold base"
    }
  ]
}
\end{lstlisting}
\caption{Example object captions generated for a nightstand scene with decorative items.}
\label{fig:caption_example}
\end{figure}

\subsubsection{Example Generated Captions}

Figure~\ref{fig:caption_example} shows example captions generated by our pipeline for a nightstand scene.

\subsection{Spatial Relationship Analysis}

To provide precise geometric context for assembly planning, we incorporate 3D bounding box data for each object, including dimensions, center positions, and spatial extents. We also generate a top-down visualization (X-Z plane projection with Y-axis flattened) showing the spatial layout of all objects with numbered labels, as shown in Figure~\ref{fig:bbox_vis}. The bounding box data is used in conjunction with visual observations in the subsequent assembly plan generation stage.

\subsection{Assembly Plan Generation}

The assembly plan generation uses a unified prompt that produces two interconnected outputs through a chain-of-thought process: (1) explicit spatial relationship analysis for each part (as described in the previous subsection), and (2) the sequential assembly plan. This two-part output structure encourages the model to first reason about geometric relationships before constructing the assembly sequence, ensuring logical consistency and physically grounded instructions.

The VLM receives multiple types of input images to understand both the complete scene context and individual object details, as illustrated in Figure~\ref{fig:assembly_inputs}: (1) the complete scene view (the one selected as front view), (2) individual object views showing each part's location in the scene (ind\_x.png), (3) red highlighted masks labeling each part (mask\_x.png), and (4) the bounding box visualization. Additionally, pre-generated captions and bounding box data for each part are provided as textual input. The complete prompt is shown in Figure~\ref{fig:assembly_prompt}.

\begin{figure*}[t]
\begin{tcolorbox}[colback=gray!10, colframe=gray!50, boxrule=0.5pt, title=Assembly Plan Generation Prompt]
\tiny
\textbf{System Instruction:} You are an indoor scene analyzer specializing in furniture assembly and spatial relationships. You will receive:
\begin{enumerate}
    \item Entire scene view with all objects together
    \item Individual views of each part in the world coordinate system
    \item Pre-generated captions for each part
    \item 3D bounding box data (dimensions, center positions, volume)
    \item Top-down bounding box visualization (X-Z plane projection)
\end{enumerate}

\textbf{Coordinate System:} +Z is forward, +X is right. Use Y differences for up/down positioning (higher Y = upper), Z differences for front/back positioning (higher Z = farther from camera).

\textbf{Your goals are to:}
\begin{enumerate}
    \item Determine the spatial relationship of each part relative to the overall group/scene
    \item Generate a construction/assembly plan for the entire scene that describes how to build it step by step
    \item Focus on relative locations and positioning of parts
    \item Use the 3D bounding box data to understand precise spatial relationships and object sizes
\end{enumerate}

\textbf{IMPORTANT:} If multiple assets are similar, be careful to distinguish them clearly by their specific spatial locations and contextual details.

\textbf{IMPORTANT:} Carefully identify whether objects have adjacent relationships (next to each other) vs support relationships (one object supporting/holding another). This distinction is crucial for understanding the scene structure and construction order.

\textbf{For each part, analyze:}
\begin{itemize}
    \item Position relative to other parts (e.g., left, right, upper, lower, front, back, center)
    \item Spatial relationships with other components
    \item Functional role in the overall furniture/scene assembly
    \item Support relationships (what supports this part, what this part supports)
    \item Physical dimensions and size relationships between objects
\end{itemize}

\textbf{For the assembly plan:}
\begin{itemize}
    \item Start with foundational/base objects (e.g., tables, floors, large furniture)
    \item Progress from large to small objects
    \item Describe the placement order that respects gravity (supporting objects before supported objects)
    \item Include precise spatial relationships and dependencies between parts
    \item Each step should describe the positioning and relative location
    \item Consider object dimensions when describing placement and clearances
\end{itemize}

\textbf{Format your response as a JSON object:}

\texttt{\{"parts": [\{"part\_idx": 0, "caption": "use provided caption", "spatial\_relation": "position relative to whole"\}, ...], "assembly\_plan": [\{"instructions": "step 1 describing positioning", "part\_idx": 2, "assembly\_idx": 0\}, ...\}]}

\textbf{CRITICAL REQUIREMENTS for parts:}
\begin{itemize}
    \item Each item must have exactly three keys: `part\_idx', `caption', `spatial\_relation'
    \item Use the provided captions as base, but supplement them if they lack support structures mentioned in assembly plan
    \item spatial\_relation should describe the part's position relative to the whole scene and other parts
\end{itemize}

\textbf{CRITICAL REQUIREMENTS for assembly\_plan:}
\begin{itemize}
    \item Each item must have exactly three keys: `instructions', `part\_idx', `assembly\_idx'
    \item part\_idx follows the idx of objects from the `parts', and each part\_idx from 0 to N-1 (where N is total parts) must appear exactly once
    \item No part\_idx should be repeated or omitted
    \item `assembly\_idx' goes from 0 to N-1
    \item Instructions should describe the placement/positioning of that specific part
    \item You can use `assembly object x' to refer to the part with assembly\_idx=x, but never use `part x' to refer objects in the assembly instructions
    \item Order the assembly steps logically (foundation first, then objects placed on them)
\end{itemize}

Keep spatial descriptions precise and actionable for assembly purposes.

\textbf{Here is an example of the expected format:}

\textbf{User Prompt:} Generate spatial relationships and assembly plan using the provided captions. Provide the JSON response as specified.
\end{tcolorbox}
\caption{Assembly plan generation prompt used to produce both spatial relationship analysis and sequential assembly instructions.}
\label{fig:assembly_prompt}
\end{figure*}

\subsubsection{Example Generated Assembly Plan}

Figure~\ref{fig:assembly_example} shows a complete example of the assembly plan generated for the nightstand scene shown in previous figures. The output demonstrates both the spatial relationship analysis (in the \texttt{parts} section) and the sequential assembly instructions (in the \texttt{assembly\_plan} section), following the format specified in the prompt.

\begin{figure*}[t]
\centering
\begin{lstlisting}[language=json, basicstyle=\tiny\ttfamily]
{
  "parts": [
    {
      "part_idx": 0,
      "caption": "A decorative potted plant featuring a dense arrangement of light blue hydrangea flowers and green leaves in a smooth, round white vase, designed to be placed on a flat surface.",
      "spatial_relation": "Positioned on the upper-left surface of the nightstand (Part 3), to the left of the table lamp (Part 2), serving as a decorative accent."
    },
    {
      "part_idx": 1,
      "caption": "A rectangular picture frame with an elegant, molded light beige finish and a white image, featuring a kickstand for freestanding tabletop display.",
      "spatial_relation": "Positioned on the upper-right surface of the nightstand (Part 3), to the right of the table lamp (Part 2), providing a personal decorative touch."
    },
    {
      "part_idx": 2,
      "caption": "A modern table lamp featuring a tapered round light-colored fabric shade with dark trim, mounted on a sturdy, angular dark brown or black square base, providing illumination.",
      "spatial_relation": "Positioned centrally on the top surface of the nightstand (Part 3), behind and between the potted plant (Part 0) and the picture frame (Part 1), serving as the primary source of illumination."
    },
    {
      "part_idx": 3,
      "caption": "A sleek, grey nightstand or cabinet featuring a flat top surface for display, two drawers with a gold-toned rectangular pull handle, and supported by a striking gold-colored metal base, providing stability and storage.",
      "spatial_relation": "The primary and foundational furniture piece, located at the base of the scene, supporting all other decorative and functional items (Part 0, Part 1, Part 2) on its top surface."
    }
  ],
  "assembly_plan": [
    {
      "instructions": "Place the sleek, grey nightstand or cabinet (assembly object 0) as the foundational element, providing the main surface and structure for the entire scene.",
      "part_idx": 3,
      "assembly_idx": 0
    },
    {
      "instructions": "Position the modern table lamp (assembly object 1) with tapered round shade centrally on the top surface of the nightstand (assembly object 0), ensuring it is stable and serves as a backdrop for smaller items.",
      "part_idx": 2,
      "assembly_idx": 1
    },
    {
      "instructions": "Place the decorative potted plant with blue hydrangea flowers (assembly object 2) on the top surface of the nightstand (assembly object 0), to the left of the table lamp (assembly object 1), enhancing the decor.",
      "part_idx": 0,
      "assembly_idx": 2
    },
    {
      "instructions": "Arrange the rectangular picture frame with light beige finish (assembly object 3) on the top surface of the nightstand (assembly object 0), to the right of the table lamp (assembly object 1), completing the decorative arrangement.",
      "part_idx": 1,
      "assembly_idx": 3
    }
  ]
}
\end{lstlisting}
\caption{Example assembly plan generated for a nightstand scene with decorative items.}
\label{fig:assembly_example}
\end{figure*}

%% file: sec/supp_impl_details.tex
\section{Implementation Details}

\subsection{Model Architecture}

Following~\citet{deng2025emerging}, our model employs a Mixture of Transformers (MoT) architecture with two specialized transformer backbones: one dedicated to scene understanding and another to scene generation. Each transformer backbone is configured with a hidden size of 3584 and consists of 28 layers. The attention mechanism employs Grouped-Query Attention (GQA)~\citep{ainslie2023gqa} with 28 query heads and 4 key-value heads.
Each transformer contains 7 billion active parameters, yielding a total of 14 billion parameters across both transformers.
The understanding transformer processes text instructions and 3D VAE latents. An extra linear layer is applied after the patchification of 3D VAE latents before sending to the transformer.
The generation transformer only works on denoising the VAE latents. Another two linear layers are applied to map 3D VAE latents to the same size as the transformer's hidden size and map back.
For positional embedding for 3D understanding and generation, we also follow~\citet{deng2025emerging} and extend it to 3D: instead of applying sinusoidal positional encoding to 2D spatial coordinates, we partition the embedding vector into three segments to encode the X, Y, and Z dimensions independently.


\subsection{\ardplus Training Details}

During training, we set the maximum sequence length to 20480 tokens. We use greedy algorithm to select sequences to form a long sequence until reach the maximum sequence length.
The clean ground truth VAE latents are used as inputs for the 3D understanding block, while noised ground truth VAE latents are used for the generation block.
To handle sequences that may exceed the maximum sequence length, we apply selective dropout strategies. For generation blocks, we randomly retain only 1 to 3 generation blocks. For refinement, we select at most 7 objects to have their refinement steps in one sequence. Additionally, we apply dropout with probability 0.1 to both 3D understanding blocks and text blocks for CFG's use.


\subsection{Inference Details}

During autoregressive generation, we maintain a key-value cache for already-generated tokens to avoid redundant computation. The KV cache is updated incrementally as new tokens are generated.
To obtain the final 3D model with appearance, we first use TRELLIS's 3D VAE decoder~\citep{trellis2024} to recover sparse structured latents from our generated tokens. We then apply TRELLIS's second-stage diffusion model, conditioned on the same text prompts of this step, to generate higher-resolution 3D Gaussian Splatting (3DGS) representations with detailed appearance and texture information.
After generating fine-grained 3D assets in object space using \ardplus, we merge them back into the scene space. We calculate the bounding box of the coarse-grained object in scene space, then force it to be a cube by fixing the center then extending its dimensions to match the longest edge.
The fine-grained object-space 3D assets are then inserted into the corresponding region of the scene-space representation according to the former bounding box.


%% file: sec/supple_eval_set.tex
\section{Training Data Details}

Our training data consists of \rev{two} complementary datasets, totaling approximately \rev{730K} samples, which provide comprehensive coverage of single objects \rev{and} complete indoor scenes.

\textbf{Single Object Dataset (\rev{500K}).}
\rev{We adopt the TRELLIS~\citep{trellis2024} single-object dataset of around 500K assets, consisting of} everyday objects across various categories including small objects, cartoon animals, characters and buildings. All objects are accompanied by descriptive captions labeled by a VLM that emphasize properties including geometry, appearance, materials, structural components, support surfaces, and functional features.

\textbf{Internal Scene Dataset (230K).} Our scene dataset comprises 230K indoor scenes, representing various room types including bedrooms, living rooms, dining rooms, home offices, and kitchens. Each scene contains multiple objects (2-15) arranged in realistic spatial configurations, along with assembly plans generated through our automated data curation pipeline.

When using single-object data for training, we also incorporate refinement steps to enable the model to learn fine-grained geometry generation. To create training diversity, we randomize the location and scale of the coarse generation step, ensuring the model learns to refine objects at various positions and sizes.

%% file: sec/supp_limitation.tex
\section{Limitations and Future Work}

While our proposed method demonstrates promising results in language-conditioned 3D scene generation, we acknowledge \rev{three} main limitations that present opportunities for future research.
\begin{itemize}
    \item Despite our explicit spatial condition, the generated layouts are not always perfect. Specifically, overlapping or gaps between two adjacent objects are not always resolved correctly. While our assembly plan provides sequential placement instructions and relative spatial relationships, the spatial condition representation (VAE tokens) may lack sufficient granularity to precisely encode fine-grained spatial positioning. Instead of relying solely on VAE tokens, we can explore more fine-grained encoding by focusing only on surface area in a higher resolution, or use other native 3D encoders~\citep{sonata} that show better spatial understanding accuracy.
    \item The size of the first placed object (especially when it is a foundational element like a sofa and table) might be too large relative to the intended scene bounds, leaving insufficient space to accommodate subsequent objects according to the assembly plan. This can lead to spatial constraint violations. To alleviate this, we can design an adaptive scene generation framework where the available spatial bounds adjust dynamically based on the scale of all objects generated up to now. This would inform the model to allocate appropriate space before placing the next object.
    \item Since our current method is based on a relatively large base model for its text understanding capacity, the generation time of a complex scene usually takes $5-10\times$ longer than existing 3D generators, like TRELLIS. Due to the pure transformer architecture, inference efficiency could be further improved using modern transformer serving platforms and infrastructure.
\end{itemize}